\documentclass[10pt,twocolumn,letterpaper]{article}

\usepackage{iccv}
\usepackage{times}
\usepackage{epsfig}
\usepackage{graphicx}
\usepackage{amsmath}
\usepackage{amssymb}

\usepackage{subfigure}
\usepackage{booktabs}
\usepackage{diagbox}
\usepackage{cite}
\usepackage{multirow}
\usepackage{enumitem}
\usepackage{color}

\usepackage[breaklinks=true,bookmarks=false]{hyperref}

\iccvfinalcopy 

\def\iccvPaperID{5052} 

\newcommand{\BG}[1]{}
\newcommand{\mysection}[1]{\vspace{3pt}\noindent\textbf{#1.}}

\ificcvfinal\fi

\newcommand{\boldparagraph}[1]{\vspace{0.5em}\noindent{\bf #1} }

\newcommand{\side}[1]{ \rotatebox{90}{#1} }

\newcommand{\nLoss}{\mathcal{L}}            
\newcommand{\nx}{\mathbf{x}}                
\newcommand{\nv}{\mathbf{v}}                
\newcommand{\nz}{\mathbf{z}}                
\newcommand{\nc}{\boldsymbol{\mu}}          

\begin{document}

\title{3D Instance Segmentation via Multi-Task Metric Learning}

\author{Jean Lahoud \\
KAUST \\
\and
Bernard Ghanem\\
KAUST\\
\and
Marc Pollefeys \\
ETH Zurich\\
\and
Martin R. Oswald \\
ETH Zurich\\
}


\maketitle
\thispagestyle{empty}

\begin{abstract}
We propose a novel method for instance label segmentation of dense 3D voxel grids\footnote{\url{https://sites.google.com/view/3d-instance-mtml}}.
We target volumetric scene representations, which have been acquired with depth sensors or multi-view stereo methods and which have been processed with semantic 3D reconstruction or scene completion methods. The main task is to learn shape information about individual object instances in order to accurately separate them, including connected and incompletely scanned objects.
We solve the 3D instance-labeling problem with a multi-task learning strategy.
The first goal is to learn an abstract feature embedding, which groups voxels with the same instance label close to each other while separating clusters with different instance labels from each other.
The second goal is to learn instance information by densely estimating directional information of the instance's center of mass for each voxel.
This is particularly useful to find instance boundaries in the clustering post-processing step, as well as, for scoring the segmentation quality for the first goal.
Both synthetic and real-world experiments demonstrate the viability and merits of our approach.
In fact, it achieves state-of-the-art performance on the ScanNet 3D instance segmentation benchmark \cite{Dai-et-al-CVPR-2017}. 
\end{abstract}


\section{Introduction}
A central goal of computer vision research is high-level scene understanding.
Recent methodological progress for 2D images makes reliable results possible for a variety of computer vision problems, including image classification \cite{Krizhevsky-et-al-NIPS-2012,Simonyan-Zisserman-ICLR-2015,Szegedy-et-al-CVPR-2015}, image segmentation \cite{Long-et-al-CVPR-2015,Ronneberger-et-al-MICCAI-2015,Badrinarayanan-et-al-TPAMI-2017}, object detection \cite{Ren-et-al-NIPS-2015,Lin-et-al-CVPR-2017,Redmon-et-al-CVPR-2016} and instance segmentation in 2D images
\cite{Dai-et-al-CVPR-2016,He-et-al-ICCV-2017,Pinheiro-et-al-NIPS-2015}.
Furthermore, it is now possible to recover highly-detailed 3D geometry with low-cost depth sensors \cite{Zach-et-al-ICCV-2007,Izadi-et-al-SIGGRAPH-2011,Niessner-et-al-SIGGRAPH-2013,Steinbruecker-et-al-ICCV-2013} or with image-based 3D reconstruction algorithms \cite{Kolev-et-al-IJCV-2009,Furukawa-Ponce-TPAMI-2010,Schoenberger-et-al-ECCV-2016}. 
Combining both these concepts, many algorithms have been developed for 3D scene and object classification \cite{Socher-et-al-NIPS-2012,Wu-et-al-CVPR-2015,Maturana-Scherer-ICIRS-2015}, 3D object detection \cite{Yang-et-al-CVPR-2018,lahoud20172d}, and joint 3D reconstruction and semantic labeling \cite{Kundu-et-al-ECCV-2014,Tateno-et-al-CVPR-2017,Dai-et-al-CVPR-2018,Dai-Niessner-ECCV-2018,Cherabier-et-al-ECCV-2018}.

%
\begin{figure}[t]
  \small
  \centering
  \newcommand{\sz}{0.4}
  \begin{tabular}{ccc}
    %
    \includegraphics[trim={1cm 1.5cm 0 0},width=0.43\columnwidth]{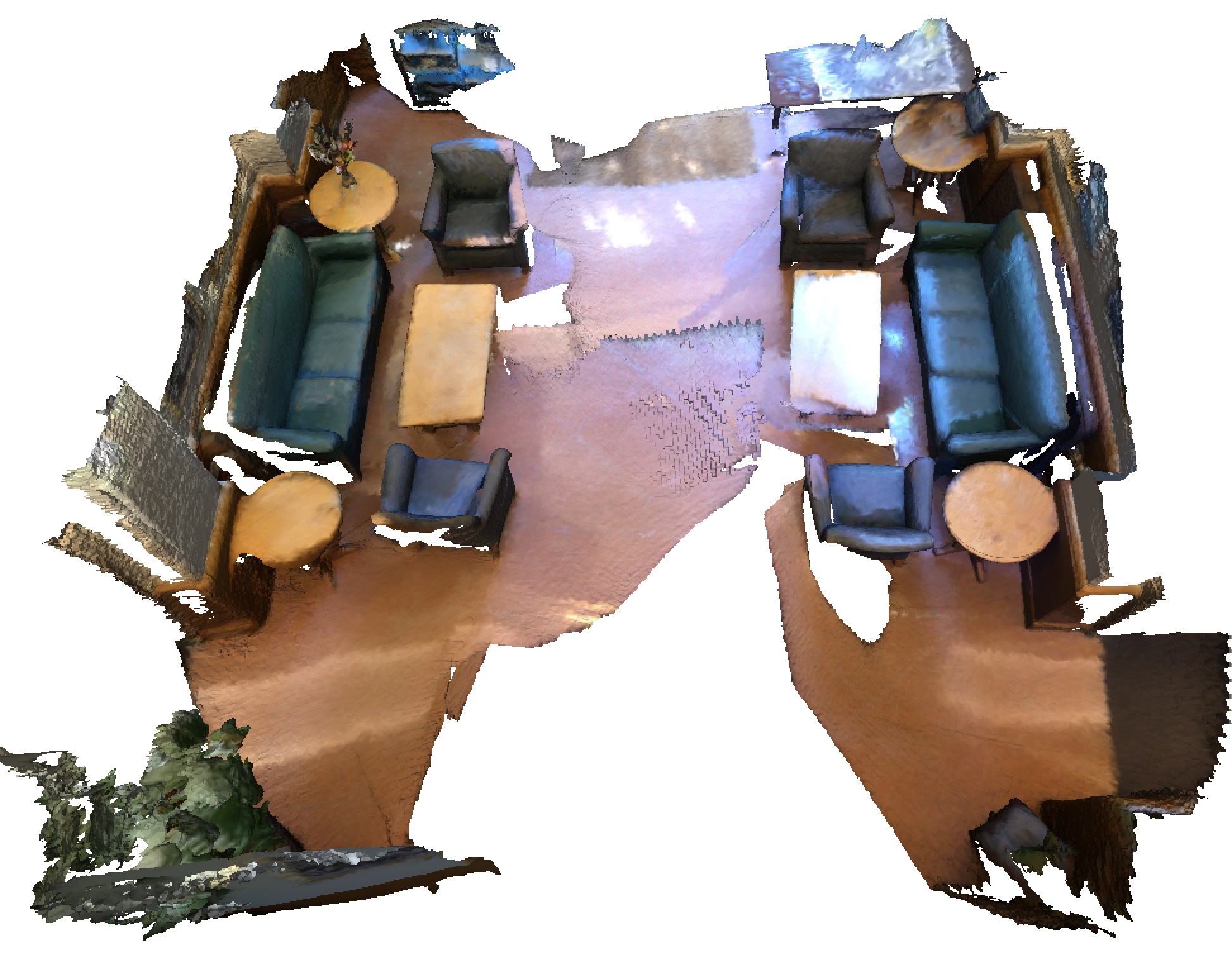} & 
    \includegraphics[trim={1cm 1.5cm 0 0},width=0.46\columnwidth]{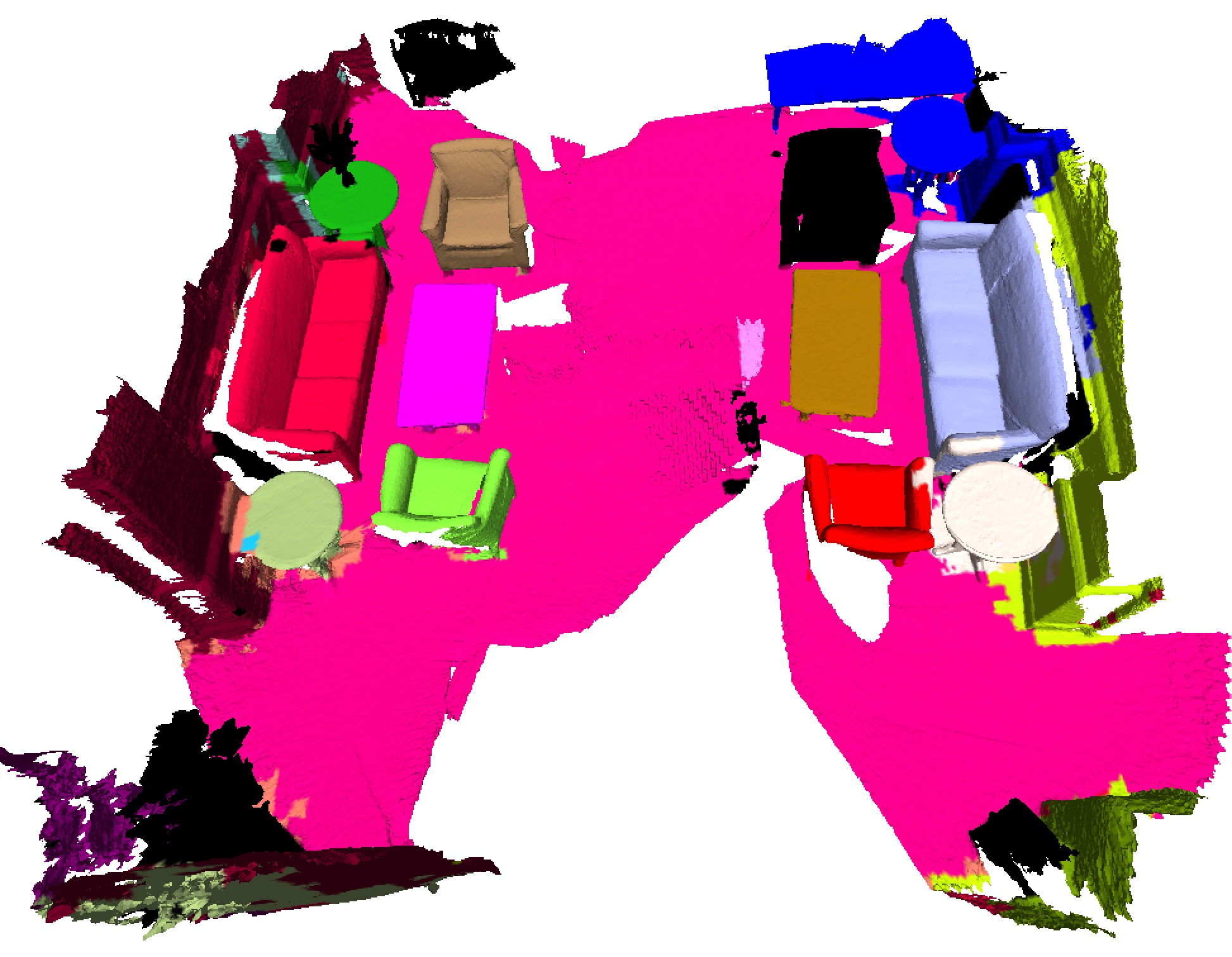} \\
    Input Scene & Our Instance Labels \\
    \includegraphics[trim={20.8cm 9cm 2cm 1cm},clip,angle=90,width=0.45\columnwidth]{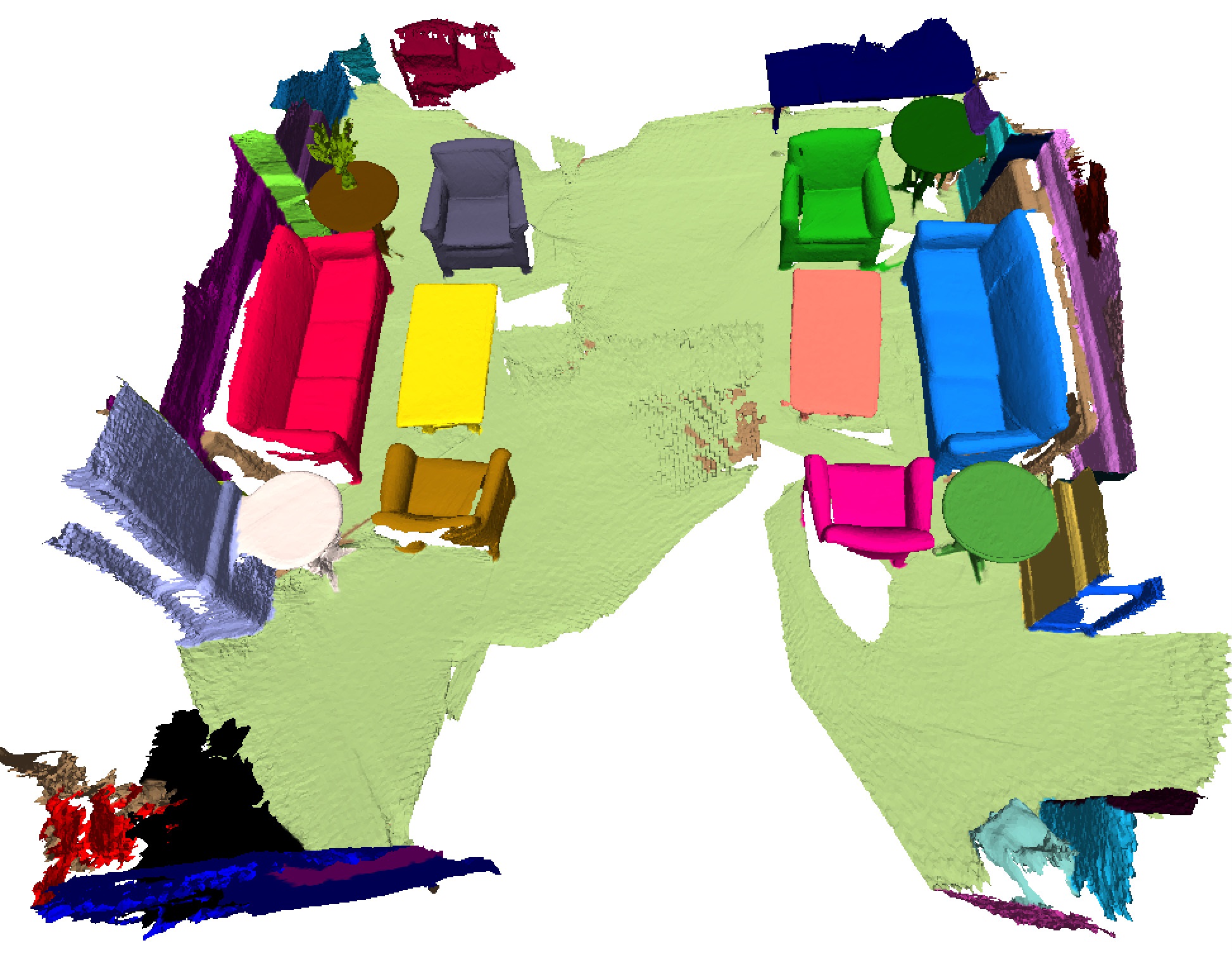} & 
    \includegraphics[trim={20.8cm 9cm 2cm 1cm},clip,angle=90,width=0.45\columnwidth]{scene0549_01_ours.jpg} \\
    Ground truth Instance Labels  & Our Instance Labels \\
  \end{tabular}
  \vspace{0.1cm}
  \caption{\textbf{Sample results of our method.} Our proposed method takes as input a 3D point cloud, and outputs instance labels unique to each object within the scene. The labels are generated by learning a metric that groups parts of the same object instance and estimates the direction towards the instance's center of mass.}
  \label{fig:teaser}
\end{figure}
%

Advances in 2D instance segmentation were mainly fueled by the large number of datasets and challenges available in the 2D realm. When compared to the plethora of powerful methods for instance segmentation of 2D images, the 3D counterpart problem has been less explored in the literature. In addition to the lack of datasets, the majority of 2D methods are not applicable to the 3D setting or their extension is by no means straightforward.

With the emergence of labeled datasets and benchmarks for the task of 3D instance segmentation (e.g. ScanNet \cite{Dai-et-al-CVPR-2017}), numerous works have surfaced to tackle this task. In many cases, the work in 3D benefits from pioneering work in 2D, with modifications that allow processing of 3D input data. As such, this 3D processing tends to be similar to other 3D understanding techniques, mainly semantic segmentation.

In this paper, we address the problem of 3D instance segmentation. Given the 3D geometry of a scene, we want to label all the geometry that belongs to the same object with a unique label. 
Unlike previous methods that entangle instance labeling with semantic labeling, we propose a technique that mainly focuses on instance labeling through grouping/clustering of information pertaining to a single object.
Our method still benefits from semantic information as a local cue, but adds to it information related to 3D dimensions and 3D connectivity, whose usefulness is unique to the 3D setting.

In particular, we propose a learning algorithm that processes a 3D voxel grid and learns two main characteristics: (1) a feature descriptor unique to every instance, and (2) a direction that would point towards the instance center. Our method aims to provide a grouping force that is independent of the size of the scene and the number of instances within. 
%
%
%
%
%
%

\mysection{Contributions} Our contributions are two fold. \emph{(i)} We propose a multi-task neural network architecture for 3D instance segmentation of voxel-based scene representations. In addition to a metric learning task, we task our network to predict directional information to the object's center. We demonstrate that the multi-task learning improves the results for both tasks. Our approach is robust and scalable, therefore suitable for processing large amounts of 3D data. \emph{(ii)} Our experiments demonstrate state-of-the-art performance for 3D instance segmentation. At the time of submission, our method ranks first in terms of average AP50 score on the ScanNet 3D instance segmentation benchmark~\cite{Dai-et-al-CVPR-2017}.

%
%
  %
  %

\section{Related Work}

This section gives a brief overview of related 2D and 3D approaches. It is worthwhile to note that a large amount of related work exists for 2D deep learning-based semantic segmentation and instance label segmentation. Recent surveys can be found in \cite{Guo-et-al-IJMIR-2017, Garcia-et-al-arXiv-2017}.

\boldparagraph{2D Instance Segmentation via Object Proposals or Detection.}
Girshick~\cite{Girshick-ICCV-2015} proposed a network architecture that creates region proposals as candidate object segments.
In a series of followup work, this idea has been extended to be faster \cite{Ren-et-al-NIPS-2015} and to additionally output pixel-accurate masks for instance segmentation \cite{He-et-al-ICCV-2017}.
The authors of YOLO~\cite{Redmon-et-al-CVPR-2016} and its follow-up work \cite{Redmon-Farhadi-CVPR-2017} apply a grid-based approach, in which each grid cell generates an object proposal. DeepMask~\cite{Pinheiro-et-al-NIPS-2015} learns to jointly estimate an object proposal and an object score. Lin~\etal\cite{Lin-et-al-CVPR-2017} propose a multi-resolution approach for object detection, which they call feature pyramid networks.
In \cite{Hayder-et-al-CVPR-2017}, the region proposals are refined with a network that predicts the distance to the boundary which is then transformed into a binary object mask.
Khoreva~\etal\cite{Khoreva-et-al-CVPR-2017} jointly perform instance and semantic segmentation.
A similar path follows \cite{Li-et-al-CVPR-2017}, which combines fully convolutional networks for semantic segmentation with instance mask proposals.
Dai~\etal\cite{Dai-et-al-CVPR-2016} use fully convolutional networks (FCNs) and split the problem into bounding box estimation, mask estimation, and object categorization and propose a multi-task cascaded network architecture.
In a follow-up work \cite{Dai-et-al-ECCV-2016}, they combine FCNs with windowed instance-sensitive score maps.

While all these approaches have been very successful in the 2D domain, many of them require large amounts of resources and their extension to the 3D domain is non-trivial and challenging.

\boldparagraph{2D Instance Segmentation via Metric Learning.}
Liang~\etal\cite{Liang-et-al-TPAMI-2017} propose a method without object proposals as they directly estimate bounding box coordinates and confidences in combination with clustering as a post-processing step. 
Fathi~\etal\cite{Fathi-et-al-arXiv-2017} compute likelihoods of pixels to belong to the same object by grouping similar pixels together within an embedding space. 
Bai and Urtasun \cite{bai2017deep} learn an energy map of the image in which object instances can be easily predicted.
Novotny~\etal\cite{Novotny-et-al-ECCV-2018} learn a position sensitive metric (semi-convolution embedding) to better distinguish between identical copies of the same object.
Kong and Fowlkes \cite{Kong-Fowlkes-CVPR-2018} train a network that assigns all pixels to a spherical embedding, in which points of the same object instance are within a close vicinity and non-instance related points are placed apart from each other.
The instances are then extracted via a variant of mean-shift clustering\cite{Fukunaga-Hostetler-TIT-1975} that is implemented as a recurrent network.
The approach by DeBrabandere~\etal\cite{DeBrabandere-et-al-arXiv-2017} follows the same idea, but the authors do not impose constraints on the shape of the embedding space.
Likewise, they compute the final segmentation via mean-shift clustering in the feature space.


None of these approaches has been applied to a 3D setting.
Our approach builds upon the work of DeBrabandere~\etal\cite{DeBrabandere-et-al-arXiv-2017}.
We extend this method with a multi-task approach for 3D instance segmentation on dense voxel grids.

\begin{figure*}[t]
	\centering
	\includegraphics[width=\textwidth]{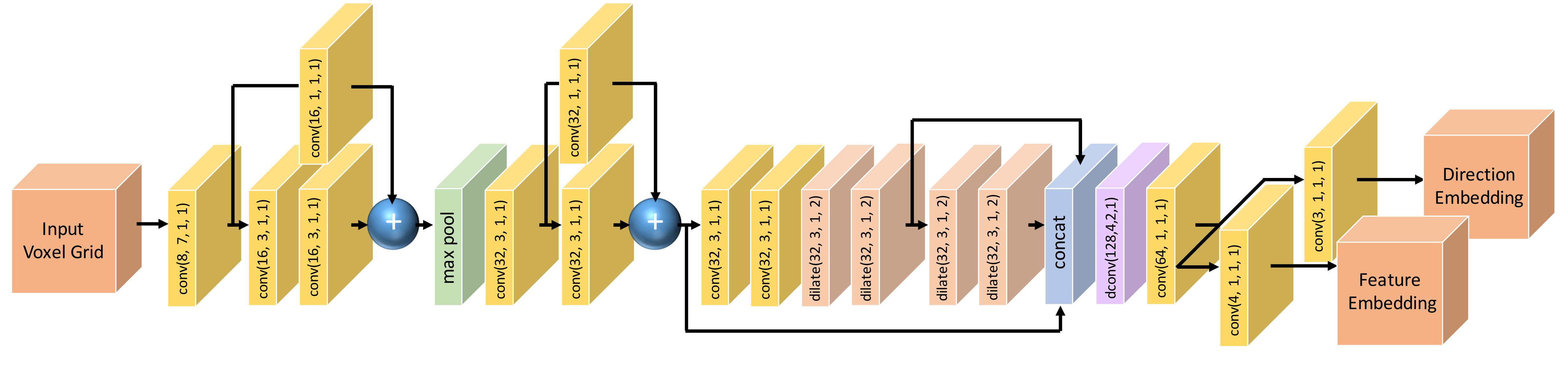}
	\vspace{-0.7cm}
	\caption{\textbf{Overview of our network architecture.} We cast 3D instance segmentation as a multi-task learning problem. The input to our method is a voxel grid and the output are two latent spaces: 1) a feature vector embedding that groups voxels with similar instance label close in the latent space; 2) a 3D latent space that encodes directional predictions for each voxel. The inputs and outputs of our network are visualized and explained in Fig.~\ref{fig:embedding_spaces}. The parameters in the figure correspond to (number of filters, kernel size, stride, dilation).}
	\label{fig:model_overview}
\end{figure*}
%


\boldparagraph{3D Instance Segmentation.}
Wang~\etal\cite{Wang-et-al-CVPR-2018} propose SGPN, an instance segmentation for 3D point clouds.
In the first step, they extract features with PointNet~\cite{Qi-et-al-CVPR-2017} and subsequently build a similarity matrix, in which each element classifies whether two points belong to the same object instance.
The approach is not very scalable and limited to small point cloud sizes, since the size of the similarity matrix is squared the number of points in the point cloud.
Moreover, there is a number of recent concurrent or unpublished works that address 3D instance segmentation. The GSPN method~\cite{yi2018gspn} proposes a generative shape proposal network, which relies on object proposals to identify instances in 3D point clouds.
The 3D-SIS approach~\cite{hou20183d} combines 2D and 3D features aggregated from multiple RGB-D input views.
MASC~\cite{liu2019masc} relies on the superior performance of the SparseConvNet~\cite{3DSemanticSegmentationWithSubmanifoldSparseConvNet} architecture and combines it with 
an instance affinity score that is estimated across multiple scales. PanopticFusion \cite{narita2019panopticfusion} predicts pixel-wise labels for RGB frames and carries them over into a 3D grid, where a fully connected CRF is used for final inference. 

Apart from these recent concurrent works, there has generally been sparse research on 3D instance segmentation.
\section{Method Overview} \label{sec:method}

In this work, we aim at segmenting 3D instances within a given 3D scene. 
To fully locate a 3D instance, one would require both a semantic label and an instance label. 
Rather than solving the complex task of scene completion, semantic labeling and instance segmentation at once, we model our 3D instance segmentation process as a post-processing step for semantic segmentation labeling.
We focus on the grouping and splitting of semantic labels, relying on inter-instance and intra-instance relations.
We benefit from the real distances in 3D scenes, where sizes and distances between objects are key to the final instance segmentation.

We split our task into a label segmentation then instance segmentation problem, as we believe that features learned in each step possess task-specific information. 
Semantic segmentation on one hand can rely on local information to predict the class label.
Learning to semantically label a volumetric representation inherently encodes features from neighboring volumes but does not require knowledge of the whole environment.
On the other hand, instance segmentation requires a holistic understanding of the scene in order to join or separate semantically labeled volumes.



\boldparagraph{Problem Setting.}
Our method's input is a voxelized 3D space with each voxel encoding either a semantic label or a local feature vector learned through semantic labeling. 
In this paper, we use the semantic labeling network in \cite{3DSemanticSegmentationWithSubmanifoldSparseConvNet}.
We fix the voxel size to preserve 3D distances among all voxels within a scene.
In problem settings where point clouds or meshes are available, one could generate a 3D voxelization by grouping information from points within every voxel.
Our method then processes the voxelized 3D space and outputs instance label masks, each corresponding to a single object in the scene, along with its semantic label.
The output mask can also be reprojected back into a point cloud by assigning the voxel label to all points within it. 


\subsection{Network Architecture} \label{subsec:network}
In order to process the 3D input, we utilize a 3D convolution network, which is based on the SSCNet architecture~\cite{Song-et-al-CVPR-2017}.
We apply some changes to the original SSCNet network to better suit our task.
As shown in Figure~\ref{fig:model_overview}, the network input and output are equally sized.
Since the pooling layer scales down the scene size, we use a transpose of convolution (also referred to as deconvolution~\cite{zeiler2010deconvolutional}) to upsample back into the original size. 
We also use larger dilations for diluted 3D convolution layers to increase the receptive field. 
We make the receptive field large enough to access all the voxels of usual indoor rooms. With a voxel size of 10cm, our receptive field is as large as 14.2m. 
With larger scenes, our 3D convolution network would still be applicable to the whole scene, while preserving the filter and voxel sizes, and thus preserving the real distances.
Objects lying at distances larger than the receptive field are separated by default. 

  


\subsection{Multi-task Loss Function} \label{subsec:loss}
In order to group voxels of the same instance, we aim to learn two types of feature embedding.
The first type maps every voxel into a feature space, where voxels of the same instance are closer to each other than voxels belonging to different instances.
This is similar to the work of DeBrabandere~\etal\cite{DeBrabandere-et-al-arXiv-2017}, but applied in a 3D setting.
The second type of feature embedding assigns a 3D vector to every voxel, where the vector would point towards the physical center of the object it belongs to.
This enables the learning of shape containment and removes ambiguities among similar shapes.

In order to learn both feature embeddings, we introduce a multi-task loss function that is minimized during training. 
The first part of the loss encourages discrimination in the feature space among multiple instances, while the second part penalizes angular deviations of vectors from the desired direction.


\begin{figure*}[t!]
  \small
  \centering  
  \includegraphics[width=0.95\textwidth]{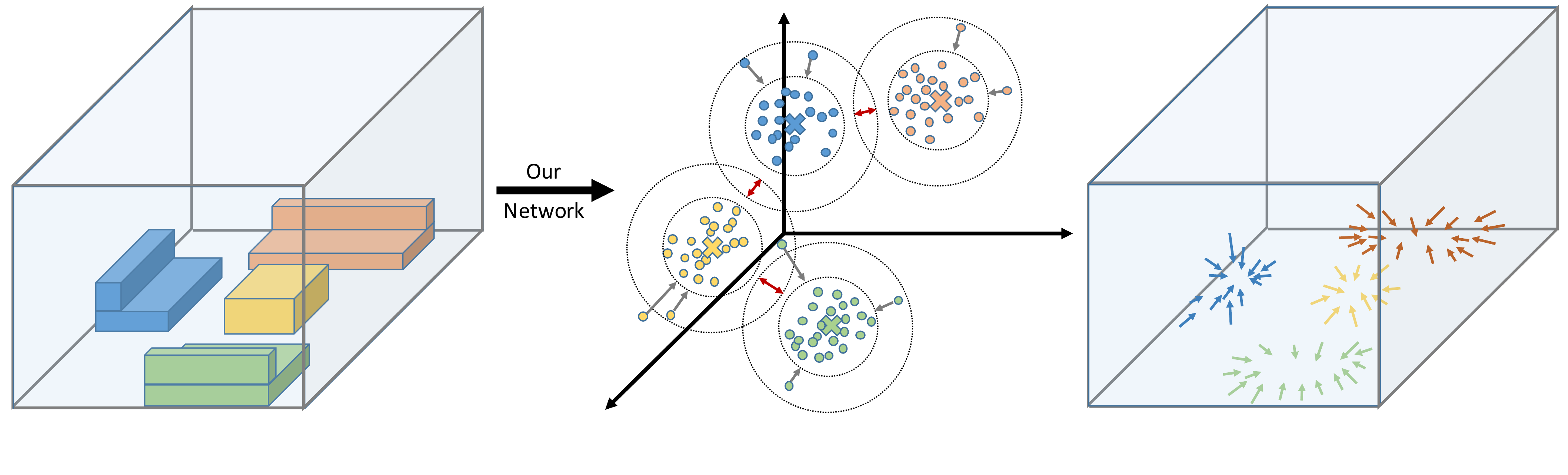} \\[-0.6cm]
  \setlength{\tabcolsep}{15mm}
  \begin{tabular}{ccc}	   	   
	   \textbf{World Space} & \hspace{0.5cm}\textbf{Feature Embedding Space} & \hspace{-1.5cm}\textbf{Direction Embedding Space}\\
	\end{tabular}
  \caption{\textbf{Embedding space visualization.}
  Voxels with similar instance labels in the world space \textbf{(left)} are mapped: (1) to similar locations in the feature embedding space such that the instances form clusters \textbf{(middle)} and (2) to directional vectors pointing to the object center \textbf{(right)}. The red arrows depict inter-class push forces among cluster centers, while the grey arrows indicate intra-class pull forces of between points and cluster centers.
  The other colors differentiate voxels or features of different object instances.
  }
  \label{fig:embedding_spaces}
\end{figure*}

\boldparagraph{Feature Embedding Loss.}
We follow the work of DeBrabandere~\etal\cite{DeBrabandere-et-al-arXiv-2017}, which learns a feature embedding that can be subsequently clustered. 
Thus, we define the feature embedding loss as a weighted sum of three terms:
\textbf{(1)} an intra-cluster variance term $\nLoss_\text{var}$ that pulls features that should belong to the same instance towards the mean feature, 
\textbf{(2)} an inter-cluster distance term $\nLoss_\text{dist}$ that encourages clusters with different instance labels to be pushed apart, and
\textbf{(3)} a regularization term $\nLoss_\text{reg}$ that pulls all features towards the origin in order to bound the activations. 
\begin{equation}
  \nLoss_\text{FE} = 
    \gamma_\text{var} \nLoss_\text{var} +
    \gamma_\text{dist} \nLoss_\text{dist} +
    \gamma_\text{reg} \nLoss_\text{reg}
  \label{eq:feature_loss}
\end{equation}
The individual loss functions are weighted by $\gamma_\text{var}=\gamma_\text{dist}=1$, $\gamma_\text{reg}=0.001$ and are defined similar to \cite{DeBrabandere-et-al-arXiv-2017} as follows:
\begin{align}
  \nLoss_\text{var} &= \frac{1}{C}\sum_{c=1}^C \frac{1}{N_c} \sum_{i=1}^{N_c}
    \left[ \|\nc_c - \nx_i\| - \delta_\text{var} \right]_{+}^2 \\
  \nLoss_\text{dist} &= \frac{1}{C(C-1)}\sum_{c_A=1}^C \sum_{\substack{c_B=1\\c_B\neq c_A}}^C
    \left[ 2\delta_\text{dist} - \|\nc_{c_A} - \nc_{c_B}\| \right]_{+}^2 \\
  \nLoss_\text{reg} &= \frac{1}{C} \sum_{c=1}^C \|\nc_c\|
\end{align}
Here $C$ is the number of ground truth clusters, $N_c$ denotes the number of elements in cluster $c$, $\nc_c$ is the cluster center, \ie the mean of the elements in cluster $c$, and $\nx_i$ is a feature vector.
Further, the norm $\|\cdot\|$ denotes the $\ell_2$-norm and $[x]_{+} = \max(0,x)$ the hinge.
The parameter $\delta_\text{var}$ describes the maximum allowed distance between a feature vector $\nx_i$ and the cluster center $\nc_c$ in order to belong to cluster $c$.
Likewise, $2\delta_\text{dist}$ is the minimum distance that different cluster centers should have in order to avoid overlap.
A visualization of the forces and the embedding spaces can be found in Figure~\ref{fig:embedding_spaces}.
Feature embeddings of different clusters exert forces on each other, \ie each feature embedding is affected by the number and location of other cluster centers.
This connection might be disadvantageous in some cases, especially when a large number of instances exist in a single scene. 
Therefore, we propose next an additional loss that provides local information essential for instance separation without being affected by other instances.

\boldparagraph{Directional Loss.} We here aim to generate a vector feature that would locally describe the intra-cluster relationship without being affected by other clusters. 
We choose the vector to be the one pointing towards the ground truth center of the object. \BG{you seem to be using object, instance, and cluster interchangeably. It's advised to be consistent with the naming}
To learn this vector feature, we attend to the following  directional loss:


\begin{equation}
  \nLoss_\text{dir} = -\frac{1}{C} \sum_{c=1}^C \frac{1}{N_c} 
                     \sum_{i=1}^{N_c} \nv_i ^{\top} \nv_i^{GT}
  \quad \text{with} \;\;
  \nv_i^{GT} = \frac{\nz_i-\nz_c}{\|\nz_i-\nz_c\|}
\end{equation}

Here, $\nv_i$ denotes the normalized directional vector feature, $\nv_i^{GT}$ is the desired direction which points towards the object center, $\nz_i$ is the voxel center location, and $\nz_c$ is the object center location. 

\boldparagraph{Joint Loss.} We jointly minimize both the feature embedding loss and the directional loss during training. Our final joint loss reads as:
\begin{equation}
   \nLoss_\text{joint} = \alpha_\text{FE}\nLoss_\text{FE} + \alpha_\text{dir}\nLoss_\text{dir}
\end{equation}
We use $\alpha_\text{FE} = 0.5$ and $\alpha_\text{dir} = 1$.

\boldparagraph{Post-processing.} 
We apply mean-shift clustering \cite{Fukunaga-Hostetler-TIT-1975} on the feature embedding. Similar to object detection algorithms, instance segmentation does not restrict the labeling to one coherent set, and thus allows overlap between multiple objects.
We use the mean-shift clustering output with multiple thresholds as proposals that are scored according to their direction feature consistency. 
We also use connected components for suggested splitting that would further be scored by the coherency of its feature embeddings. The coherency of the feature embedding is described by the number of feature embeddings that lie within a given threshold from the feature cluster center. The directional feature coherency score is simply $\nLoss_\text{dir}$, which is the average  cosine similarity between the normalized vector pointing from the voxel to the center of the object and the predicted normalized direction feature. We then sort all object proposals and perform non-maximum suppression (NMS) to remove objects that overlap by more than a threshold.
The final score is obtained by appending both feature embedding scores with a score that encourages objects of regular sizes over extremely large or small objects.
As for the semantic label, it is chosen to be the most occurring label among all points within the clustered voxels.

\subsection{Network Training} \label{subsec:training}

\boldparagraph{Training Data.} During training, we append flips of voxelized scenes as well as multiple orientations around the vertical axis to our training data.
We pretrain our network using ground truth segmentation labels as input, with labels one-hot encoded to maintain the same sized input as training using the semantic segmentation output. 

\section{Results and Evaluation}

\boldparagraph{Setup.}
Our network was implemented in Tensorflow and run with an Nvidia GTX1080Ti GPU.
For the network training, we use the ADAM optimizer and a learning rate of $5\mathrm{e}{-4}$ and batch size of $2$. 
The training converged after about $100$ epochs and took about $2$ days. The inference time for our network is about $1$s for scene sizes of $1.6$M voxels.

\begin{figure}[tb]
	\centering
	\includegraphics[width=\columnwidth]{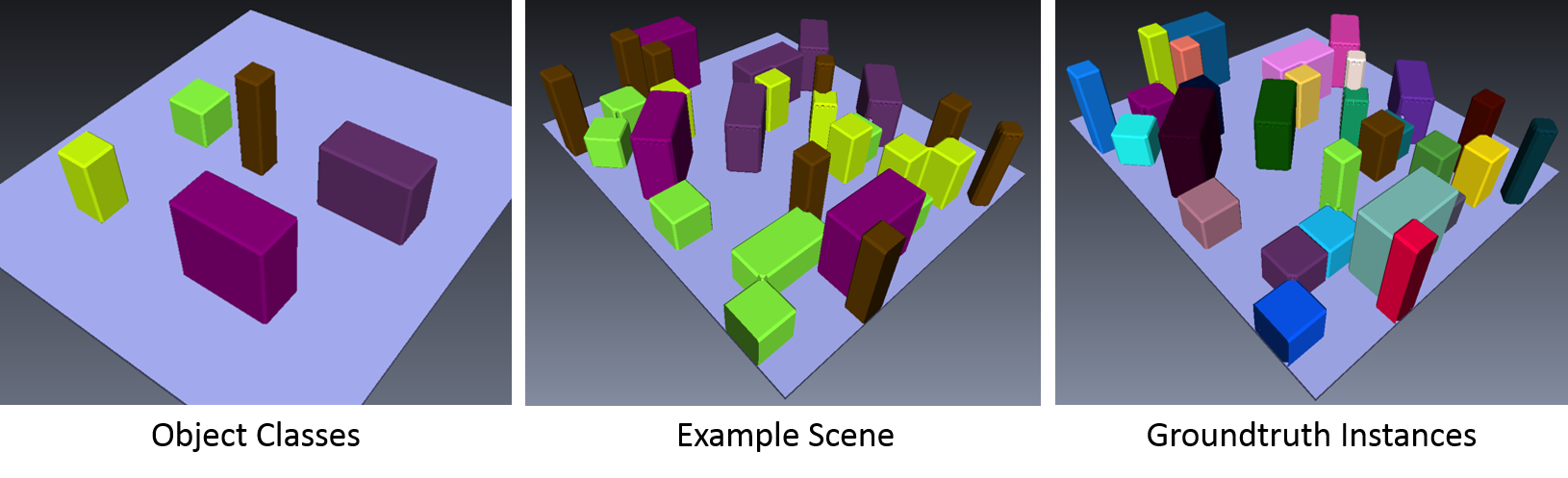}
	\caption{\textbf{Overview of the synthetic toy dataset.}
	\textbf{Left:} We consider 5 different object classes represented by cubes with various edge lengths. 
	\textbf{Middle:} Example scene with object colors showing the class labels.
	\textbf{Right:} Corresponding ground truth instance labeling (randomly chosen color per instance).
	}
	\label{fig:synthetic_gt}
\end{figure}

\boldparagraph{Datasets.}
For experimental evaluation, we trained and tested our method on the following datasets that include real and synthetic data.
\begin{itemize}[topsep=2pt,leftmargin=*]
\setlength\itemsep{0mm}
  \item \textbf{Synthetic Toy Dataset:} In order to validate our approach, we create a synthetic dataset with objects of different sizes and aspect ratios placed on a planar surface.
  We introduce 5 object shapes, where each shape is analogous to an object class in the real data.
  The shapes of the objects considered are shown in Figure~\ref{fig:synthetic_gt}.
  We then randomly orient and position objects on the surface plane, and randomly choose whether an object is in contact with another object.
  We generate 1000 scene, and split our dataset into 900 training scenes, and 100 testing scenes.
  \item \textbf{ScanNet~\cite{Dai-et-al-CVPR-2017}:} We conduct experiments on the ScanNet v2 dataset, which contains 1513 scans with 3D instance annotations.
  The training set contains 1201 scans, and the remaining 312 scans are used for validation.
  An additional 100 unlabeled scans form an evaluation test set. 
  
\end{itemize}
%

\boldparagraph{Evaluation metrics.} Following the evaluation procedure adopted in most instance segmentation methods as well as the ScanNet evaluation benchmark, we use the average precision metric (AP) score to evaluate our proposed algorithm. 
We use the AP25 and AP50 metrics, which denote the AP score with a minimum intersection-over-union (IoU) threshold of 25\% and 50\%, respectively. 
The AP score averages scores obtained with IoU thresholds ranging from 50\% to 95\% with a step of 5\%.

\boldparagraph{Baselines.}
To assess the performance of our method, we consider the following baseline methods:
%
%
\begin{itemize}[topsep=2pt,leftmargin=*]
\setlength\itemsep{0mm}
\item \textbf{Input Segmentation:} In this case, we assume that the segmentation label, which is input to our method, to be the desired instance segmentation label. If every scene contains a single instance of every semantic label, this baseline would be ideal. In reality, these scenes barely occur, but such a metric would still serve as an inception to whether splitting and/or grouping voxels is reasonable.  
\item \textbf{Connected Components:} Given the ground truth segmentation labels, a connected components algorithm tends to correctly label all instances that are not touching. 
Since this happens seldom in a 3D setting, this is usually a high-scoring and challenging baseline.
\item We further compare against submissions to the ScanNet benchmark, specifically \textbf{MaskRCNN proj}~\cite{He-et-al-ICCV-2017}, \textbf{SGPN}~\cite{Qi-et-al-CVPR-2017}, \textbf{GSPN}~\cite{yi2018gspn}, \textbf{3D-SIS}~\cite{hou20183d}, \textbf{Occipital-SCS}, \textbf{MASC}~\cite{liu2019masc}, \textbf{PanopticFusion}~\cite{narita2019panopticfusion}, and \textbf{3D-BoNet}~\cite{yang2019learning}.
\end{itemize}
%

\subsection{Evaluation on Synthetic 3D Data}


We evaluate our method on the simple toy dataset, and report AP50 score for all objects in Table~\ref{tab:results_synth}.
In this part, we allow only one coherent labeling.
Note that the directional loss alone is not discriminative enough for subsequent clustering and is thus not considered in the ablation study.
Generating object proposals from directional information only is tedious, since it is noisy and the clustering problem is much more difficult and less efficient. 
Therefore, we do not evaluate the directional prediction alone, but instead, we resort to using object proposals from mean shift clustering and using the directional information for scoring them.

\begin{figure}[tb]
	\centering
	
	\includegraphics[width = \columnwidth]{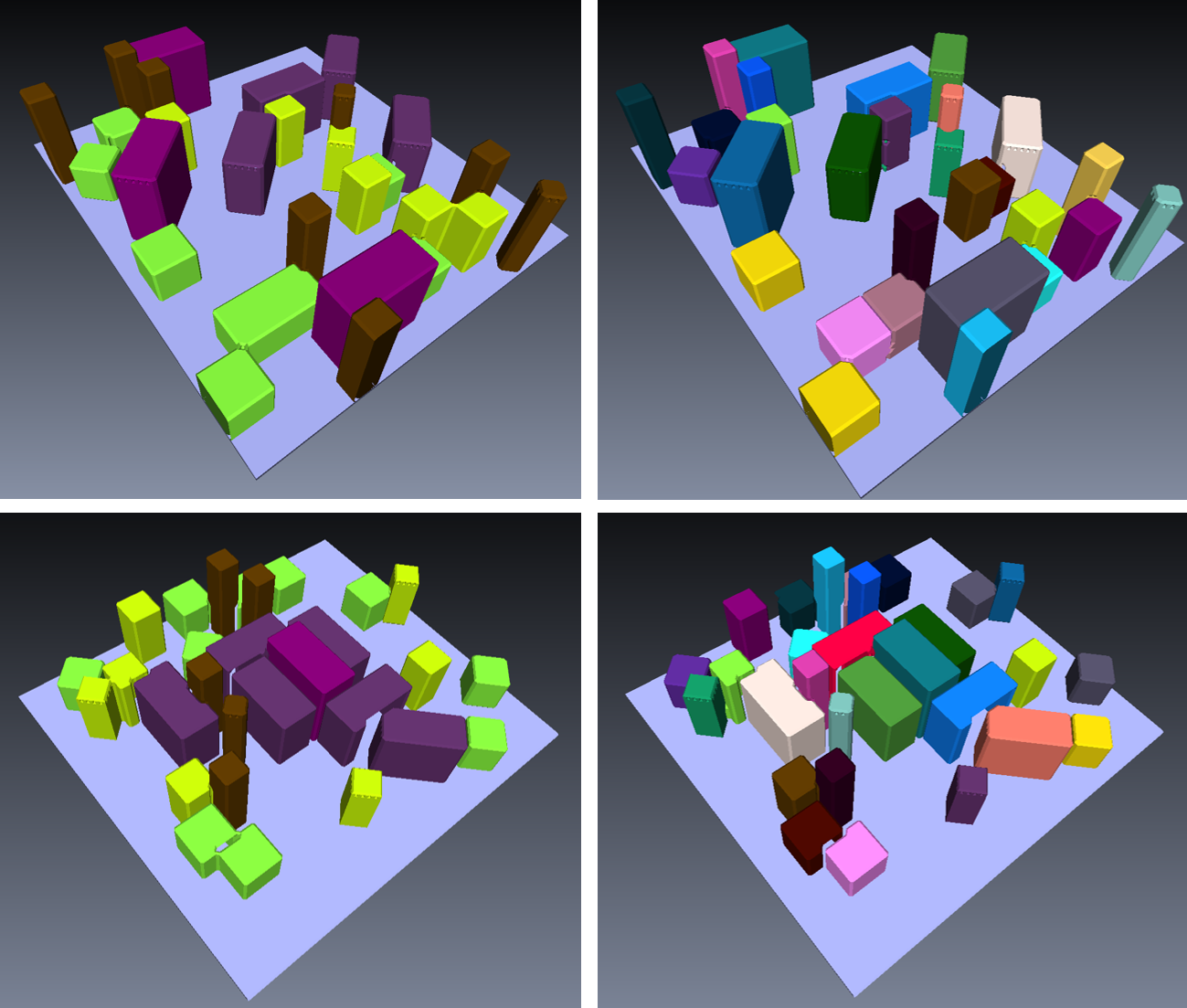}
	\scriptsize
	\setlength{\tabcolsep}{4mm}
	\begin{tabular}{cc}	   	   
	   Input Scenes with Semantic Labels  & Output Scenes with Instance Labels
	\end{tabular}
	\caption{\textbf{Experiment on synthetic toy dataset.} Two examples of random scenes for which our network generated instance labels.}
	\label{fig:synthetic_output}
	\vspace{5pt}
\end{figure}

The goal of the simple toy problem in Figure~\ref{fig:synthetic_output} is to study whether the network can abstract and differentiate various object sizes although their shapes are rather similar. 
Furthermore, it is interesting to see how our method performs when object instances are spatially touching, especially when they belong to the same semantic class. 
Although the input features are very similar (due to the same object class and the spatial proximity), our network is able to successfully place the corresponding feature vectors in different locations in the feature space.
\begin{table}[tb!]
\centering
\vspace{6pt}
\begin{tabular}{l|ccccc}
\hline
Method & Obj1 & Obj2 & Obj3 & Obj4 & Obj5\\
\hline
\hline
Connected comp. & 92.5 & 85.1 & 86.9 & 93.5 & 79.9 \\
Ours (FE only) & 97.3 & 92.7 & 95.0 & 96.4 & 95.2\\
Ours (Multi-task) & \textbf{98.0} & \textbf{93.5} & \textbf{96.1} & \textbf{96.6} & \textbf{95.3}\\
\hline
\end{tabular}
\vspace{0.01cm}
\caption{\textbf{AP50 results on synthetic toy dataset.} On this dataset with 5 objects, our approach with multi-task learning as well as the baseline with only feature embedding (FE) outperform the connected components baseline, even though it uses the ground truth semantic labels. The difference between \textit{FE only} and \textit{Multi-task} is small in a noise-free setting.}
\label{tab:results_synth}
\vspace{5pt}
\end{table}

\subsection{Evaluation on Real 3D Data}

\begin{figure*}[ht!]
  \small
  \centering
  \vspace{-0.2cm}
  \newcommand{\sz}{0.36}
  \begin{tabular}{ccccc}
    \includegraphics[width=\sz\columnwidth]{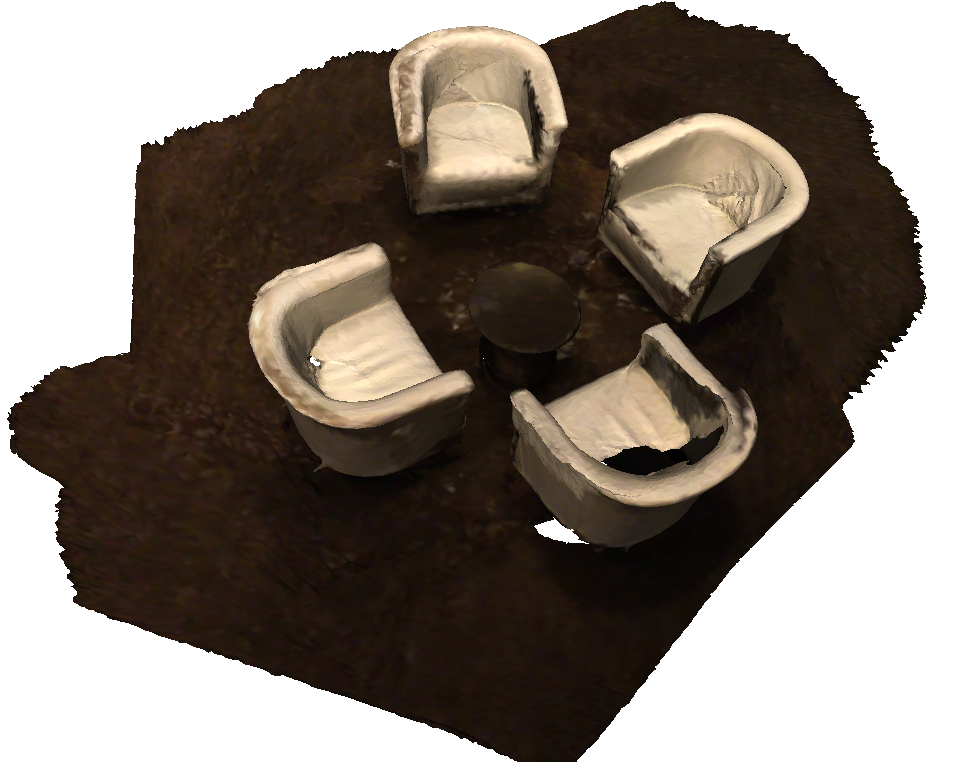} &     
    \includegraphics[width=\sz\columnwidth]{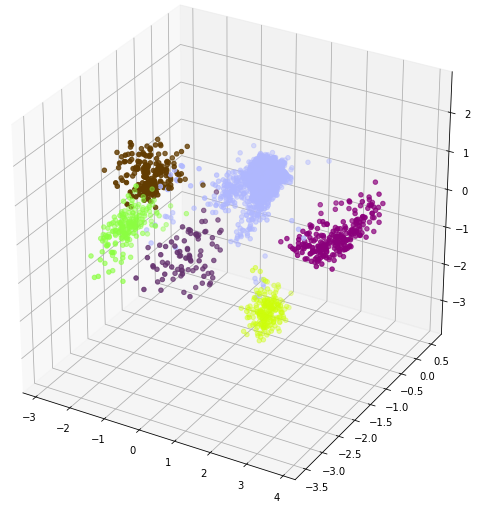} &
    \includegraphics[width=\sz\columnwidth]{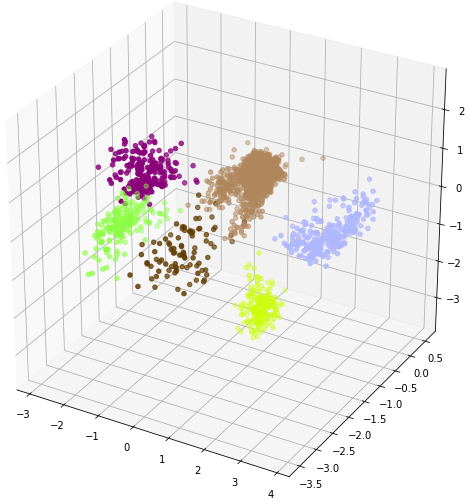} &     
    \includegraphics[width=\sz\columnwidth]{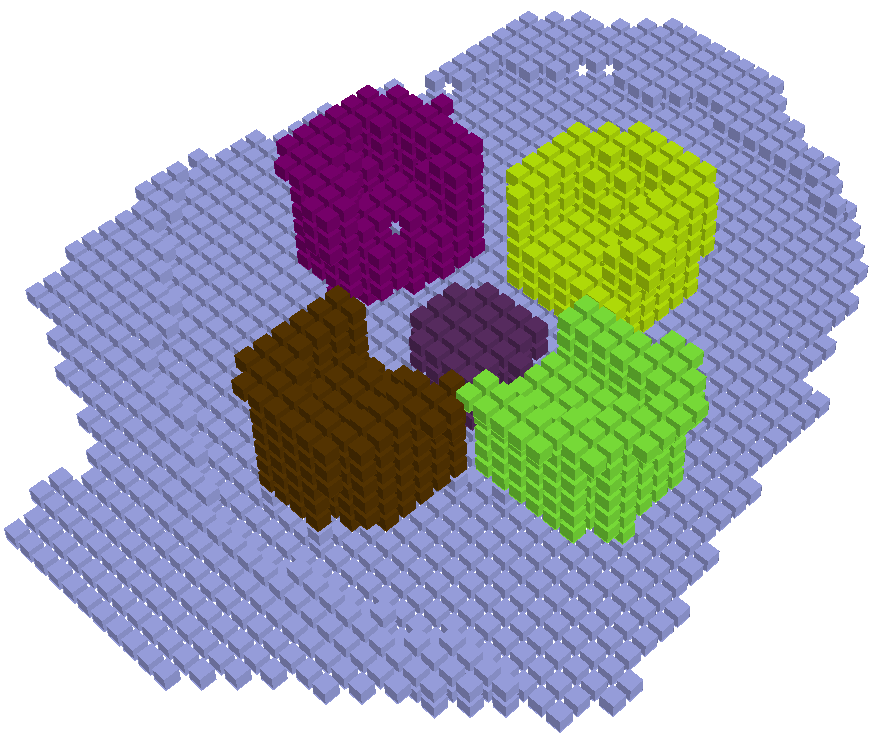} &  
    \includegraphics[width=\sz\columnwidth]{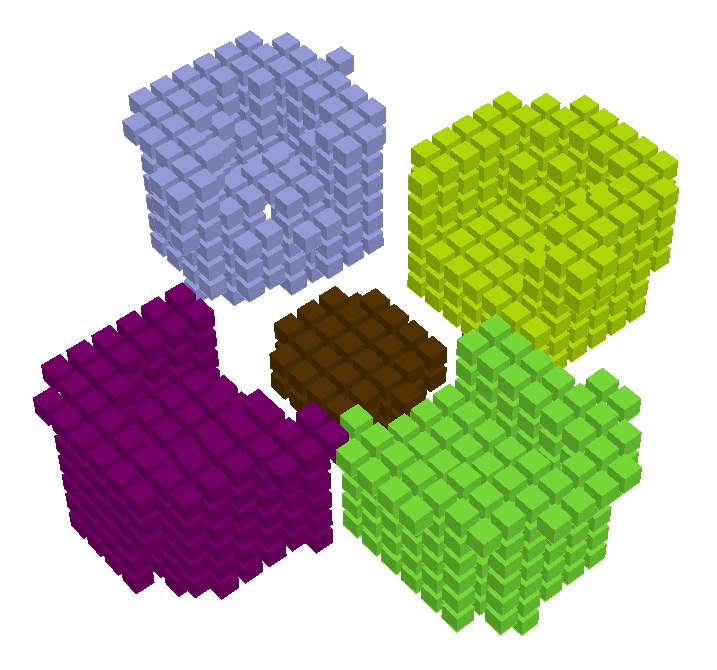} \\
    Input (RGB) & Feature Label GT & Feature Label Ours & GT Label & Clustering Label \\
  \end{tabular}
  \vspace{0.1cm}
  \caption{\textbf{Visualization of the feature embedding and labeling.} This figure shows (from left to right) the colored 3D scene input, its generated 3D feature embeddings, along with the ground truth (GT) labels and our instance labeling result after mean-shift clustering (colors of the instances in the final results are chosen randomly and do not correspond to GT label colors).}
  \vspace{-0.1cm}
  \label{fig:feature_embedding}
\end{figure*}

\boldparagraph{Feature Space Study.}
Minimizing the feature loss in Eq.~\eqref{eq:feature_loss} works toward two tasks: pulling points belonging to the same instance together and pushing clusters of different instances apart. Since real data contains noise, outliers, and missing data, the mapping of individual points in the feature space might be less discriminative and clusters might be overlapping.
In Figure~\ref{fig:feature_embedding}, we visualize the 3D feature space in order to study these effects and observe that feature points of the same instance do indeed spread towards neighboring clusters. But for this example, the feature clustering results are not influenced and still achieve high accuracy. Note that we exclude ground and wall labels since their instance segmentation and splitting is less meaningful and is also ignored in the benchmark.
%





\boldparagraph{Evaluation on ScanNet Output.}
%
%
%
%
%
%
In Figure~\ref{fig:qualitative_results_a}, we present qualitative results on the ScanNet dataset~\cite{Dai-et-al-CVPR-2017}. The results of our method on the voxel grid are simply projected onto the mesh which is then used for evaluation on the benchmark. As can be seen in the rightmost column, our method sometimes splits objects like `desk' or the labels of `furniture' bleed into neighboring geometry. Due to our mostly geometric approach, our method needs  structural changes to recognize object boundaries and to potentially relabel a new instance. Nonetheless, our proposed method was able to group single object instances together in most cases.

In Table~\ref{tab:results_val_real}, we provide an ablation study and include comparisons against simple baselines. The first baseline uses the input segmentation labels (SparseConvNet~\cite{3DSemanticSegmentationWithSubmanifoldSparseConvNet}) as instance labels. 
Furthermore, we evaluate a simple connected component labeling method on the segmentation labeling, because in the 3D setting in general, and considering the given datasets, very few object instances are touching each other.
Hence, this connected component baseline is already a challenging one especially for a rather noise free geometry and labeling. It is clear that this method tends to substantially improve the instance labeling results. With increasing amounts of noise connected component labeling rapidly performs worse. In rare cases, the results of this method get worse, which is due to the fact that the scenes are not completely scanned and a single object instance might be disconnected due to missing scene parts.

\boldparagraph{Ablation Study: Single-task vs. Multi-task.} We compare our network with single-task learning to that of multi-task learning.
%
The six rightmost columns in Table~\ref{tab:results_val_real} show the results of single-task learning and multi-task learning.
With very few exceptions, the network trained with a multi-task loss consistently outperforms the single-task one.
This is in line with the results on the synthetic dataset and supports our hypothesis that the directional loss adds more discriminative features, which are helpful to group the features according to object instances in the feature space. 
For objects that rarely have multiple instances within a scene, such as the `counter' class, the segmentation as instance outperforms our method. Since this occurrence is uncommon, its effect on the overall average evaluation is negligible.

\begin{table*}[ht!]
\centering
\setlength{\tabcolsep}{4mm}
\resizebox{\textwidth}{!}{
\begin{tabular}{l|ccc|ccc|ccc|ccc}
\hline
      & \multicolumn{3}{c|}{Segment.~\cite{3DSemanticSegmentationWithSubmanifoldSparseConvNet} as Instance} &  \multicolumn{3}{c|}{Connect. Comp. on \cite{3DSemanticSegmentationWithSubmanifoldSparseConvNet}} & \multicolumn{3}{c|}{Ours (FE only)} & \multicolumn{3}{c}{Ours (Multi-task)}\\
Class & AP & AP50 & AP25 & AP & AP50 & AP25 & AP & AP50 & AP25 & AP & AP50 & AP25\\
\hline
cabinet        &         0.002  &         0.008  &         0.039  &         0.024  &         0.081  &         0.153  &         0.036  &         0.118  & \textbf{0.396} & \textbf{0.042} & \textbf{0.145} &         0.346  \\
bed            &         0.105  &         0.197  &         0.540  & \textbf{0.200} &         0.467  &         0.651  &         0.154  &         0.446  &         0.696  &         0.197  & \textbf{0.540} & \textbf{0.806} \\
chair          &         0.000  &         0.001  &         0.027  &         0.138  &         0.239  &         0.434  &         0.475  &         0.689  &         0.814  & \textbf{0.567} & \textbf{0.792} & \textbf{0.877} \\
sofa           &         0.066  &         0.240  &         0.462  &         0.157  &         0.398  &         0.533  &         0.172  &         0.369  &         0.684  & \textbf{0.226} & \textbf{0.488} & \textbf{0.803} \\
table          &         0.027  &         0.061  &         0.160  &         0.154  &         0.324  &         0.428  &         0.207  &         0.361  &         0.593  & \textbf{0.242} & \textbf{0.427} & \textbf{0.674} \\
door           &         0.019  &         0.037  &         0.070  &         0.041  &         0.073  &         0.108  &         0.142  &         0.304  &         0.429  & \textbf{0.152} & \textbf{0.324} & \textbf{0.458} \\
window         &         0.015  &         0.023  &         0.023  &         0.020  &         0.031  &         0.037  &         0.113  &         0.258  &         0.423  & \textbf{0.152} & \textbf{0.327} & \textbf{0.472} \\
bookshelf      &         0.013  &         0.024  &         0.187  &         0.077  &         0.198  & \textbf{0.453} &         0.075  &         0.175  &         0.423  & \textbf{0.080} & \textbf{0.219} & \textbf{0.453} \\
picture        &         0.001  &         0.005  &         0.005  &         0.001  &         0.005  &         0.008  &         0.028  &         0.067  &         0.169  & \textbf{0.044} & \textbf{0.109} & \textbf{0.198} \\
counter        &         0.007  &         0.032  &         0.216  & \textbf{0.008} & \textbf{0.034} & \textbf{0.266} &         0.001  &         0.004  &         0.094  &         0.001  &         0.008  &         0.097  \\
desk           &         0.012  &         0.057  &         0.211  &         0.022  &         0.109  &         0.364  &         0.011  &         0.053  &         0.327  & \textbf{0.031} & \textbf{0.142} & \textbf{0.499} \\
curtain        &         0.034  &         0.085  &         0.185  &         0.081  &         0.173  &         0.225  &         0.114  &         0.285  &         0.450  & \textbf{0.174} & \textbf{0.399} & \textbf{0.542} \\
refrigerator   &         0.059  &         0.112  &         0.211  &         0.105  &         0.162  &         0.225  &         0.124  &         0.302  &         0.317  & \textbf{0.185} & \textbf{0.421} & \textbf{0.441} \\
shower curtain &         0.119  &         0.231  &         0.231  &         0.128  &         0.227  &         0.284  &         0.392  &         0.593  &         0.710  & \textbf{0.402} & \textbf{0.643} & \textbf{0.749} \\
toilet         &         0.326  &         0.676  &         0.701  &         0.575  &         0.801  &         0.801  & \textbf{0.636} &         0.962  &         0.977  &         0.625  & \textbf{0.965} & \textbf{0.980} \\
sink           &         0.048  &         0.130  &         0.328  &         0.054  &         0.135  &         0.307  &         0.094  &         0.294  &         0.397  & \textbf{0.120} & \textbf{0.364} & \textbf{0.445} \\
bathtub        &         0.357  &         0.677  &         0.677  & \textbf{0.319} &         0.631  &         0.700  &         0.235  &         0.553  &         0.674  &         0.311  & \textbf{0.708} & \textbf{0.794} \\
otherfurniture &         0.004  &         0.010  &         0.039  &         0.021  &         0.052  &         0.107  &         0.061  &         0.154  &         0.283  & \textbf{0.097} & \textbf{0.215} & \textbf{0.335} \\
\hline
average        &         0.068  &         0.145  &         0.239  &         0.118  &         0.230  &         0.338  &         0.171  &         0.333  &         0.492  & \textbf{0.203} & \textbf{0.402} & \textbf{0.554} \\
\hline
\end{tabular}
} 
\vspace{0.01cm}
\caption{\textbf{Ablation study on the ScanNet dataset~\cite{Dai-et-al-CVPR-2017} validation set.} 
We show the instance labeling performance of the segmentation method in \cite{3DSemanticSegmentationWithSubmanifoldSparseConvNet}, connected components labeling on the \cite{3DSemanticSegmentationWithSubmanifoldSparseConvNet} segmentation, our method with feature embedding (FE) only and our method with multi-task learning.
}
\label{tab:results_val_real}
\end{table*}


\begin{table*}[ht!]
\centering
\resizebox{\textwidth}{!}{
\begin{tabular}{l|c|cccccccccccccccccc}
\hline
Method & Avg AP &\side{bathtub}&\side{bed}&\side{bookshelf}&\side{cabinet}&\side{chair}&\side{counter}&\side{curtain}&\side{desk}&\side{door}&\side{otherfurniture}&\side{picture}&\side{refrigerator}&\side{shower curtain}&\side{sink}&\side{sofa}&\side{table}&\side{toilet}&\side{window}\\
\hline
\textbf{MTML (Ours)}      & \textbf{0.55} & \textbf{1.00} & \textbf{0.81} &         0.59  &         0.33  & \textbf{0.65} &         0.00  & \textbf{0.82} &         0.18  & \textbf{0.42} &         0.36  &         0.18  &         0.45  & \textbf{1.00} &         0.44  &         0.69  &         0.57  & \textbf{1.00} &         0.40 \\
Occipital-SCS             &         0.51  & \textbf{1.00} &         0.72  &         0.51  & \textbf{0.51} &         0.61  &         0.09  &         0.60  &         0.18  &         0.35  &         0.38  &         0.17  &         0.44  &                   0.85  &         0.39  &         0.62  &         0.54  &         0.89  &         0.39  \\
3D-BoNet                  &            0.49  & \textbf{1.00} &         0.67  &         0.59  &         0.30  &         0.48  &         0.10  &          0.62  & \textbf{0.31} &         0.34  &         0.26  &         0.13  &         0.43  &                  0.80  &         0.40  &         0.50  &         0.51  &         0.91  & \textbf{0.44} \\
PanopticFusion\cite{narita2019panopticfusion} &         0.48  &         0.67  &         0.71  & \textbf{0.60} &         0.26  &         0.55  &         0.00  &       0.61  &         0.18  &         0.25  & \textbf{0.43} & \textbf{0.44} &         0.41  &         0.86  & \textbf{0.49} &         0.59  &         0.27  &         0.94  &         0.36  \\
ResNet-backbone\cite{liang20193d}&         0.46  & \textbf{1.00} &      0.74  &         0.16  &         0.26  &         0.59  & \textbf{0.14} &         0.48  &         0.22  &        0.42 &         0.41  &         0.13  &         0.32  &         0.71  &         0.41  &         0.54  & \textbf{0.59} &         0.87  &         0.30  \\
MASC\cite{liu2019masc}           &         0.45  &         0.53  &         0.56  &         0.38  &      0.38  &         0.63  &         0.00  &         0.51  &         0.26  &         0.36  &         0.43  &         0.33  & \textbf{0.45} &         0.57  &         0.37  &         0.64  &         0.39  &         0.98 &         0.28  \\
3D-SIS\cite{hou20183d}         &         0.38  & \textbf{1.00} &         0.43  &         0.25  &         0.19  &         0.58  &         0.01  &         0.26  &         0.03  &         0.32  &         0.24  &         0.08  &         0.42  &         0.86  &         0.12  & \textbf{0.70} &         0.27  &         0.88  &         0.24  \\
Unet-backbone\cite{liang20193d}  &         0.32  &         0.67  &         0.72  &         0.23  &         0.19  &         0.48  &         0.01  &         0.22  &         0.07  &         0.20  &         0.17  &         0.11  &         0.12  &         0.44  &         0.2  &         0.62  &         0.36  &         0.92  &         0.09  \\
R-PointNet\cite{yi2018gspn}     &         0.31  &         0.50  &         0.41  &         0.31  &         0.35  &         0.59  &         0.05  &         0.07  &         0.13  &         0.28  &         0.29  &         0.03  &         0.22  &         0.21  &         0.33  &         0.40  &         0.28  &         0.82  &         0.25  \\
3D-BEVIS       &         0.25 &         0.67  &         0.57  &         0.08  &         0.04  &         0.39  &         0.03  &         0.04  &         0.10  &         0.10  &         0.03  &         0.03  &         0.10  &         0.38  &         0.13  &         0.60  &         0.18  &         0.85  &         0.17  \\
Seg-Cluster    &         0.22  &         0.37  &         0.34  &         0.29  &         0.11  &         0.33  &         0.03  &         0.28  &         0.09  &         0.11  &         0.11  &         0.01  &         0.08  &         0.32  &         0.11  &         0.31  &         0.30  &         0.59  &         0.12  \\
SGPN\cite{Wang-et-al-CVPR-2018}           &         0.14  &         0.21  &         0.39  &         0.17  &         0.07  &         0.28  &         0.03  &         0.07  &         0.00  &         0.09  &         0.04  &         0.02  &         0.03  &         0.00  &         0.11  &         0.35  &         0.17  &         0.44  &         0.14  \\
MaskRCNN proj  &         0.06  &         0.33  &         0.00  &         0.00  &         0.05  &         0.00  &         0.00  &         0.02  &         0.00  &         0.05  &         0.02  &         0.24  &         0.07  &         0.00  &         0.01  &         0.11  &         0.02  &         0.11   &         0.01 \\
\hline
\end{tabular}
} 
\vspace{0.01cm}
\caption{\textbf{State-of-the-art comparison on the ScanNet 3D instance segmentation dataset~\cite{Dai-et-al-CVPR-2017}.} The table shows the AP50 score of individual semantic categories and the average score (sorted by avg AP50 score in descending order). We achieve the best average score.} 
\label{tab:results_test}
\end{table*}

\begin{figure*}[ht!]
  \small
  \centering
  \setlength{\tabcolsep}{0.4mm}
  \newcommand{\sz}{0.14}
  \newcommand{\lsp}{0mm}
  \begin{tabular}{ccccccc}
    \includegraphics[width=\sz\textwidth]{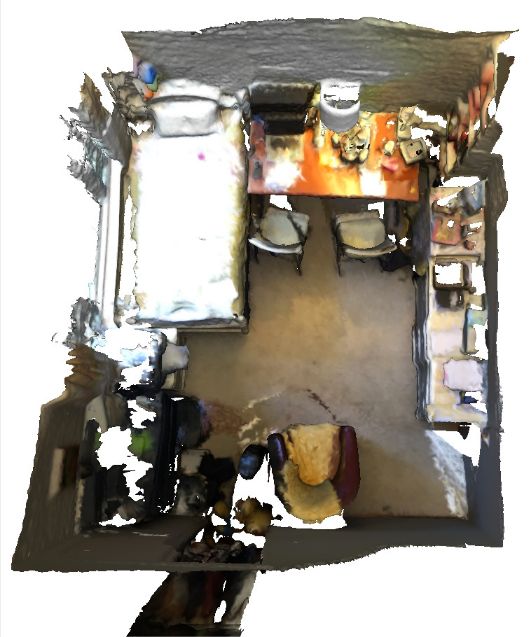} &
    \includegraphics[width=\sz\textwidth]{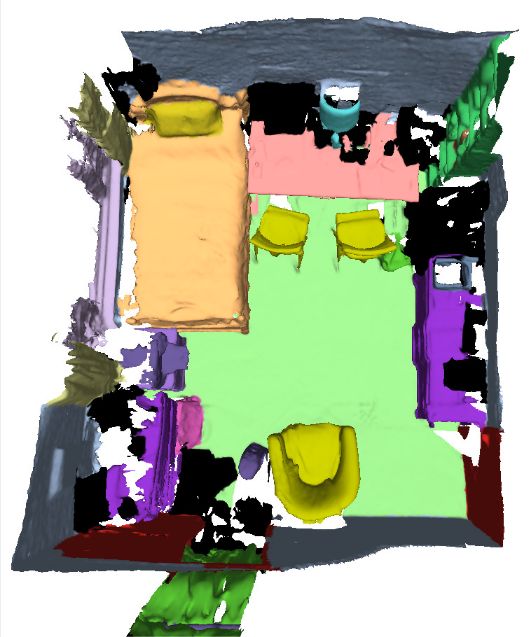} & 
    \includegraphics[width=\sz\textwidth]{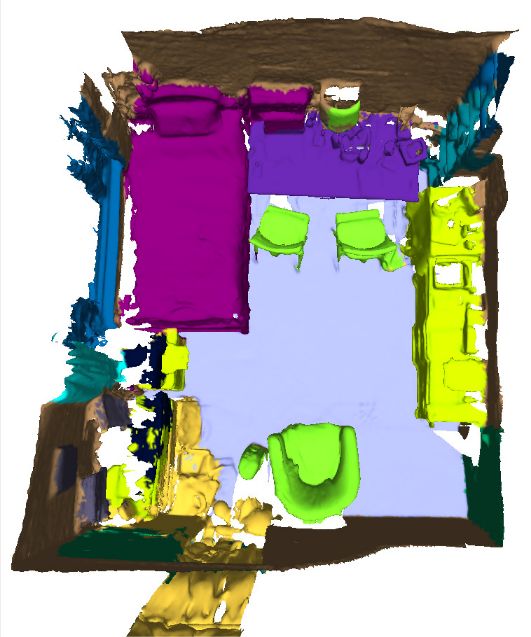} &
    \includegraphics[width=\sz\textwidth]{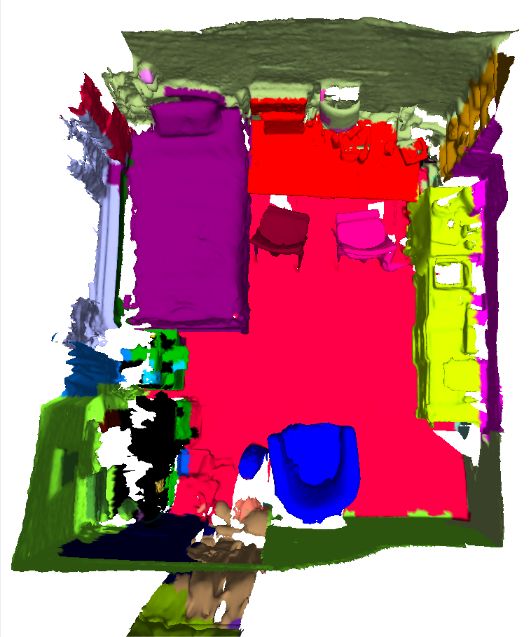} & 
    \includegraphics[width=\sz\textwidth]{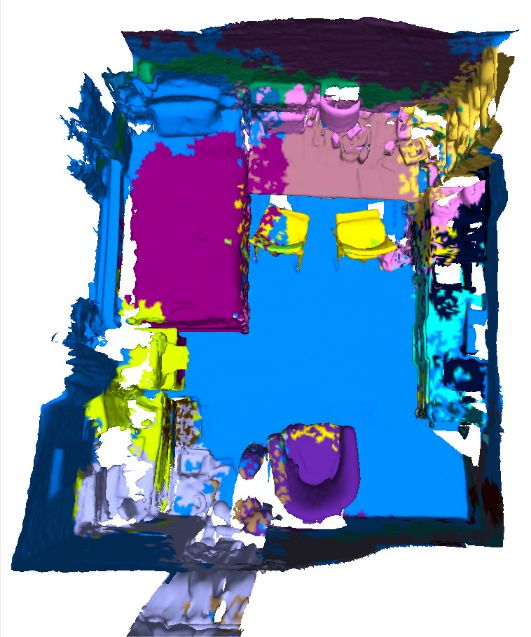} & 
    \includegraphics[width=\sz\textwidth]{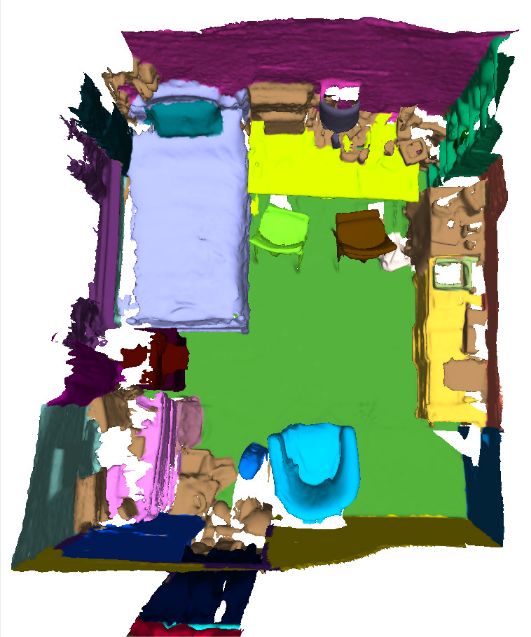} & 
    \includegraphics[width=\sz\textwidth]{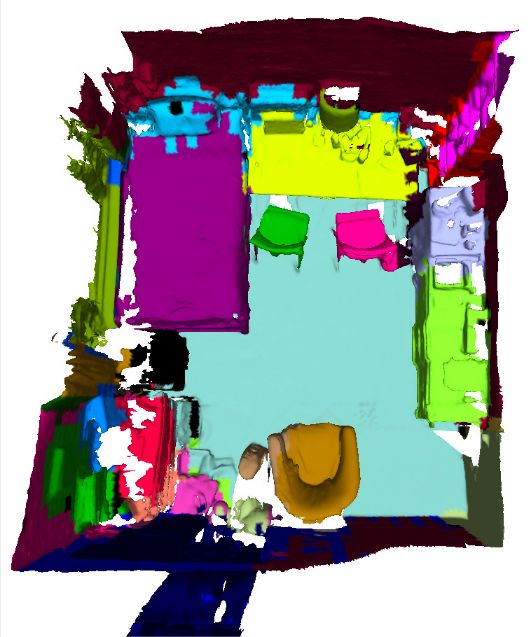} \\[\lsp]
    \includegraphics[width=\sz\textwidth]{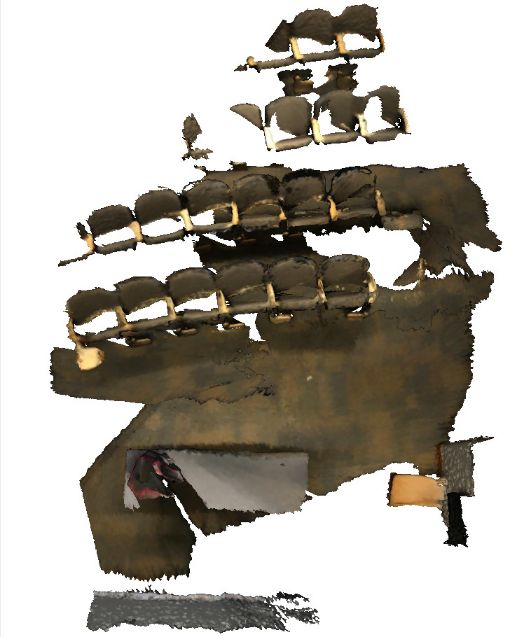} &
    \includegraphics[width=\sz\textwidth]{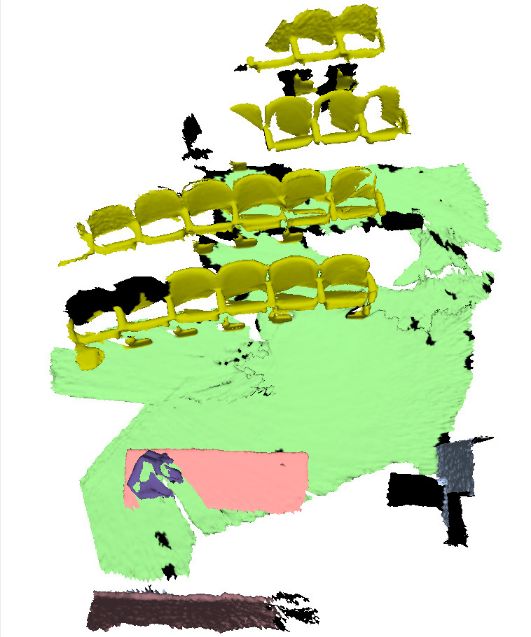} & 
    \includegraphics[width=\sz\textwidth]{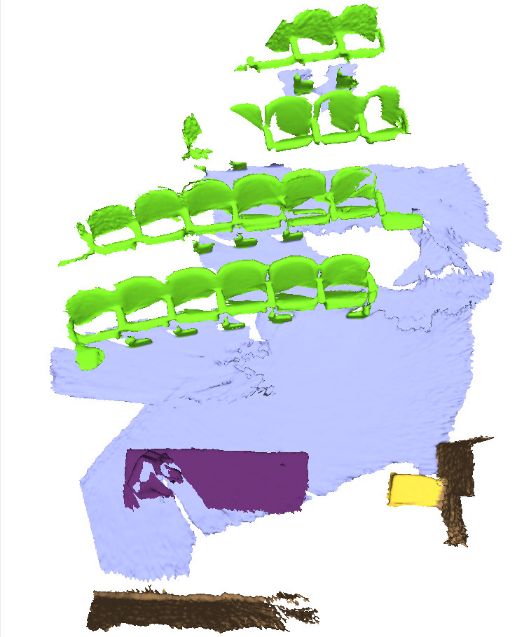} &
    \includegraphics[width=\sz\textwidth]{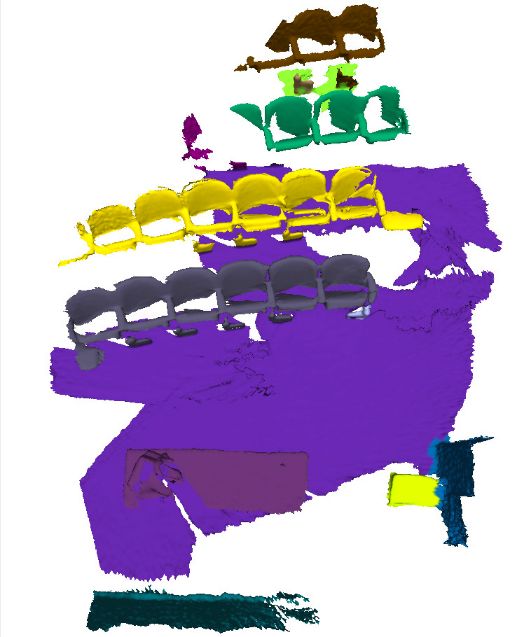} & 
    \includegraphics[width=\sz\textwidth]{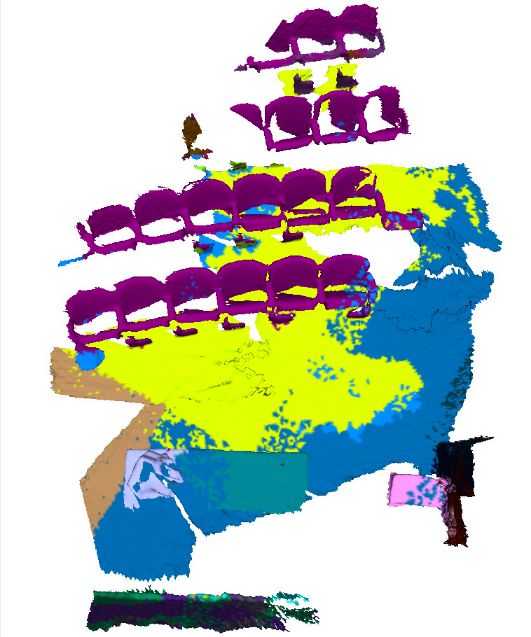} & 
    \includegraphics[width=\sz\textwidth]{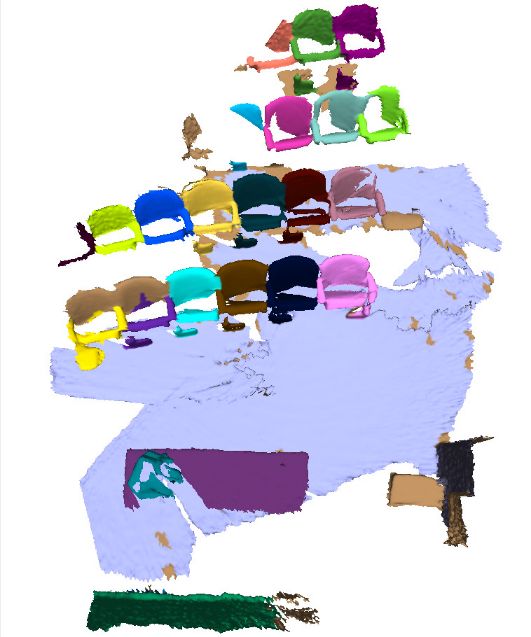} & 
    \includegraphics[width=\sz\textwidth]{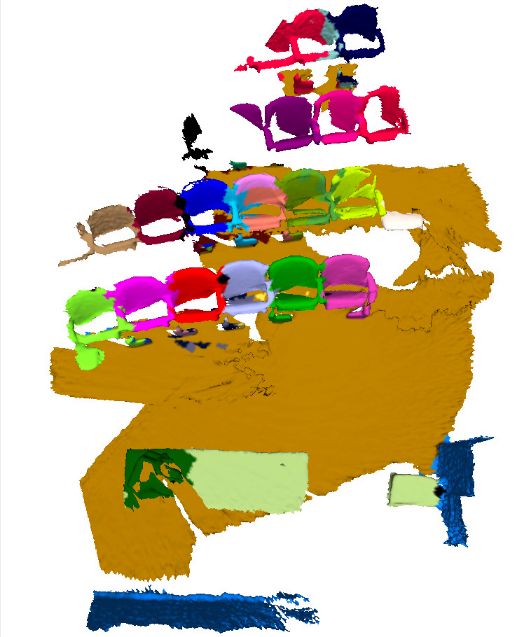} \\[\lsp]
    \includegraphics[width=\sz\textwidth]{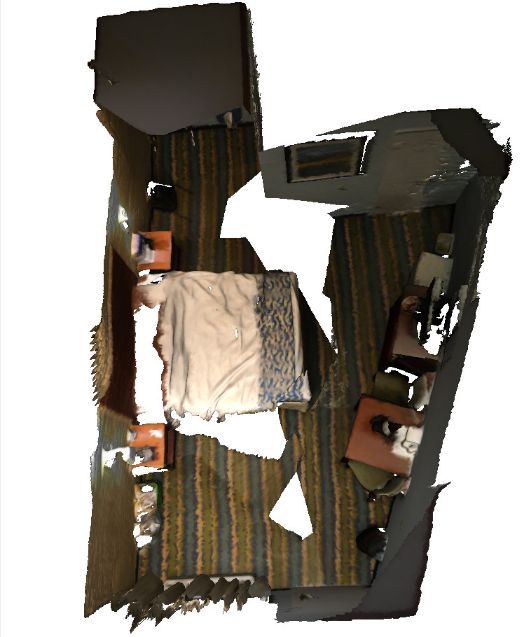} &
    \includegraphics[width=\sz\textwidth]{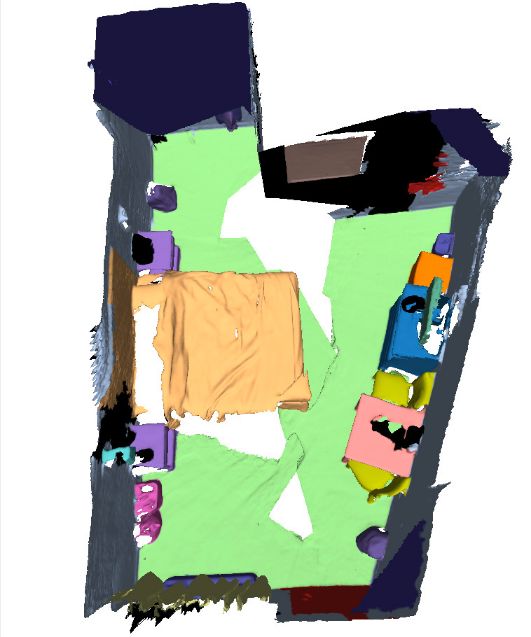} & 
    \includegraphics[width=\sz\textwidth]{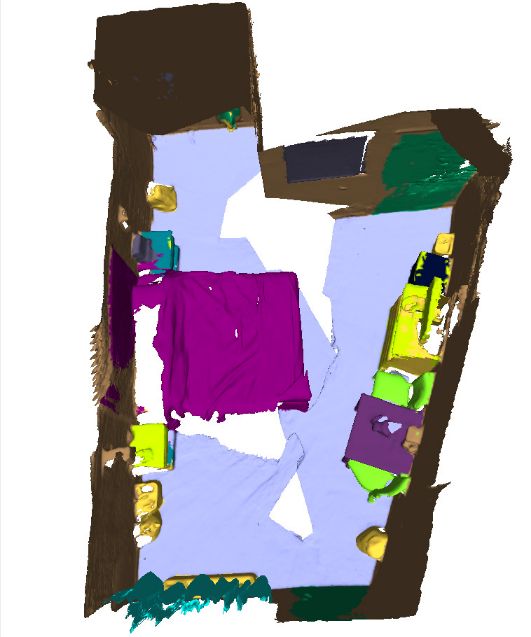} &
    \includegraphics[width=\sz\textwidth]{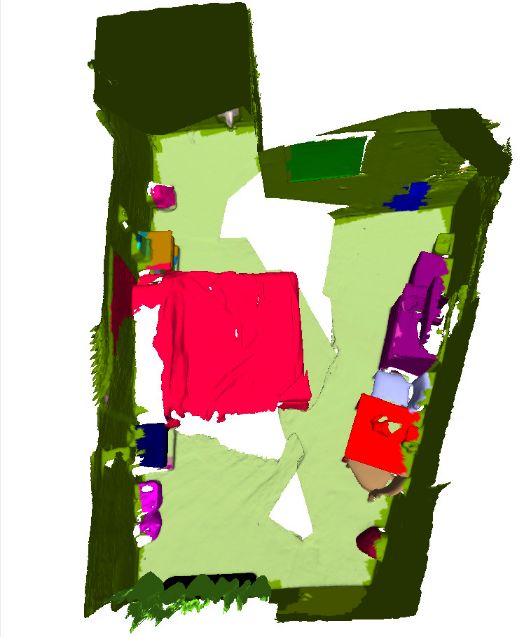} & 
    \includegraphics[width=\sz\textwidth]{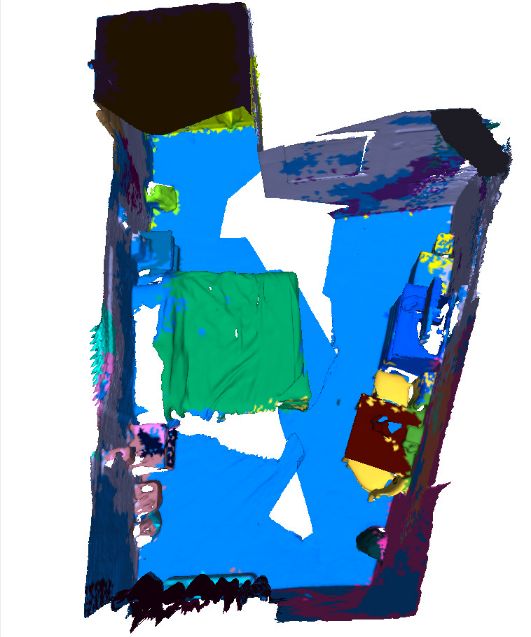} & 
    \includegraphics[width=\sz\textwidth]{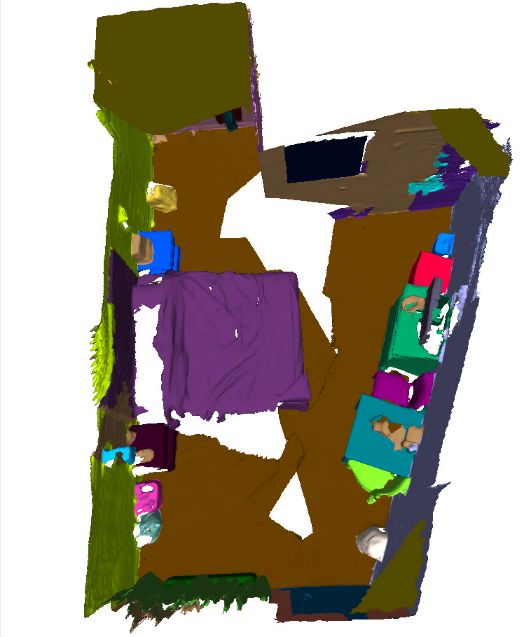} & 
    \includegraphics[width=\sz\textwidth]{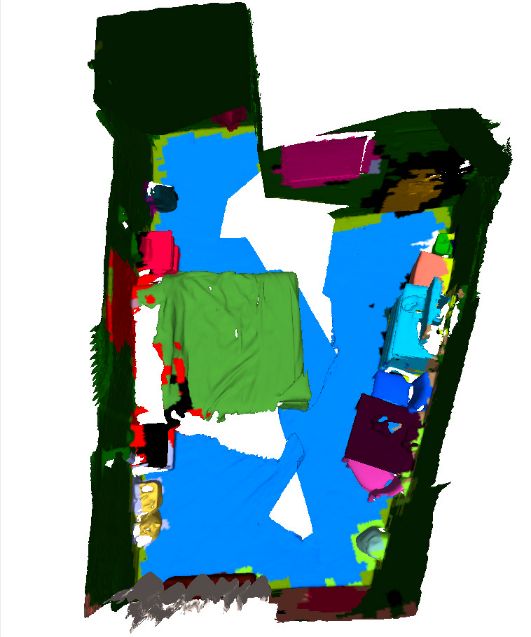} \\[\lsp]
    \includegraphics[width=\sz\textwidth]{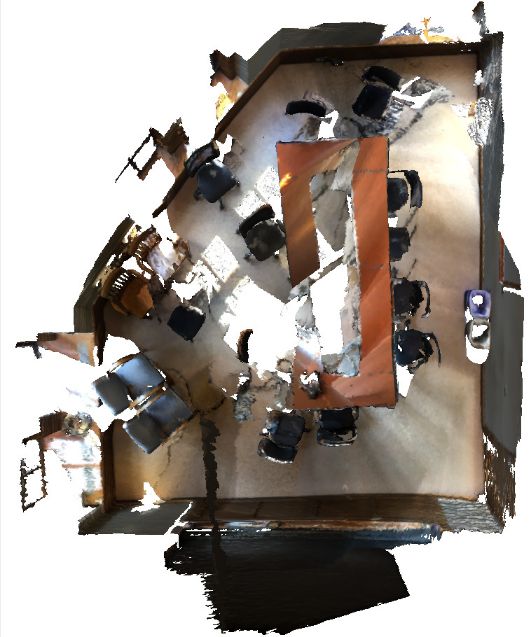} &
    \includegraphics[width=\sz\textwidth]{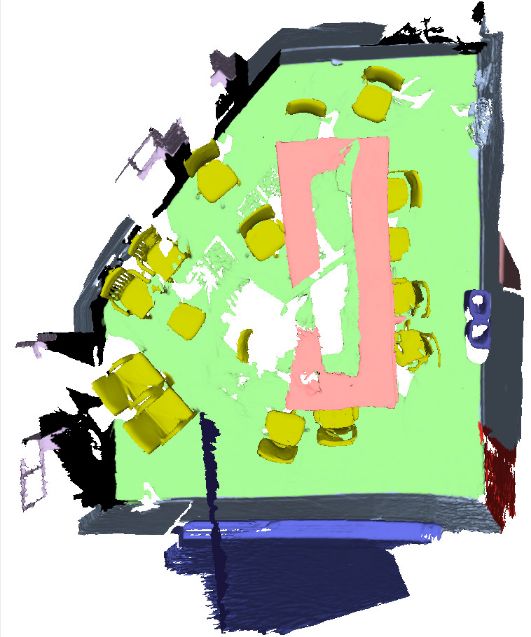} & 
    \includegraphics[width=\sz\textwidth]{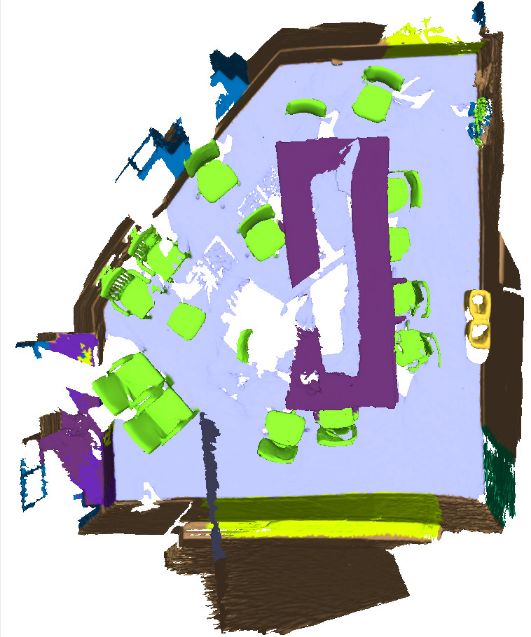} &
    \includegraphics[width=\sz\textwidth]{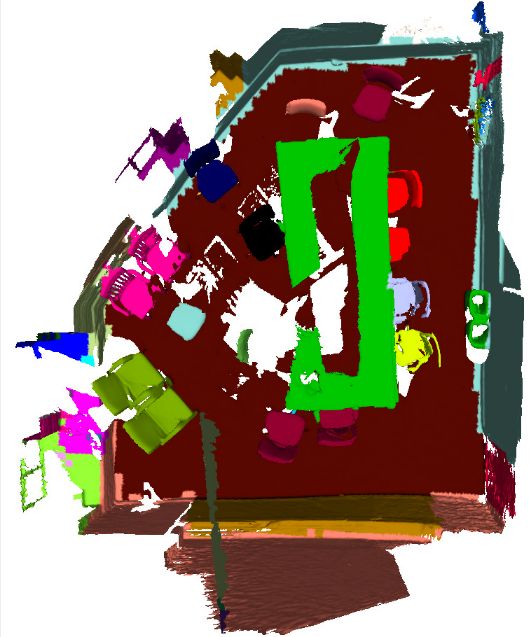} & 
    \includegraphics[width=\sz\textwidth]{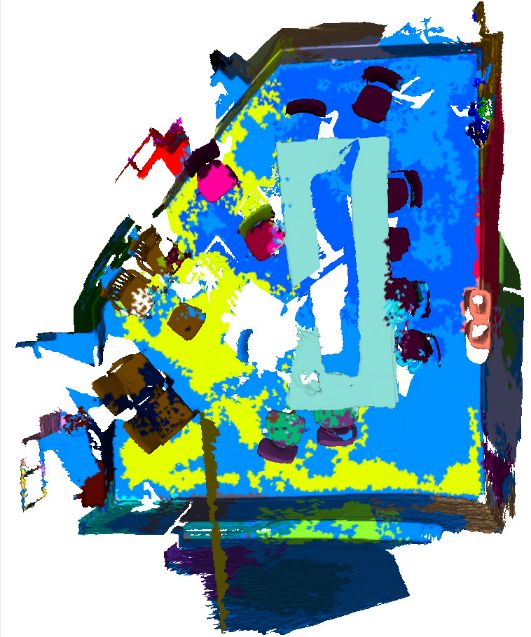} & 
    \includegraphics[width=\sz\textwidth]{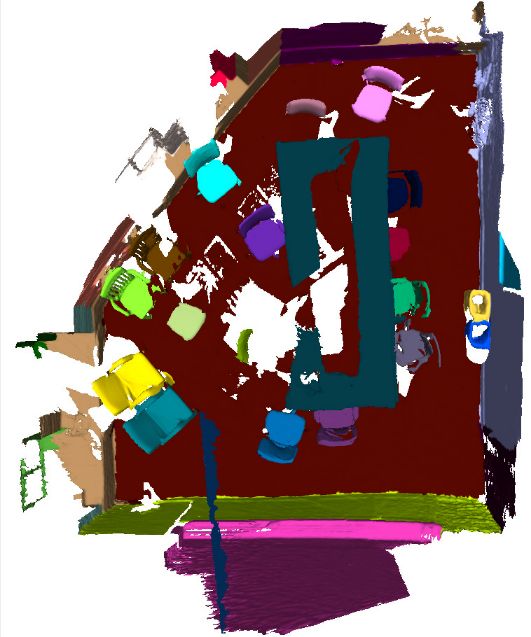} & 
    \includegraphics[width=\sz\textwidth]{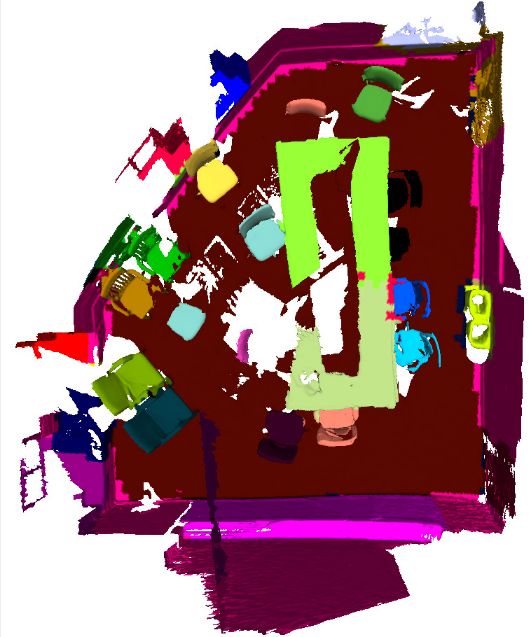} \\[\lsp]
    \includegraphics[width=\sz\textwidth]{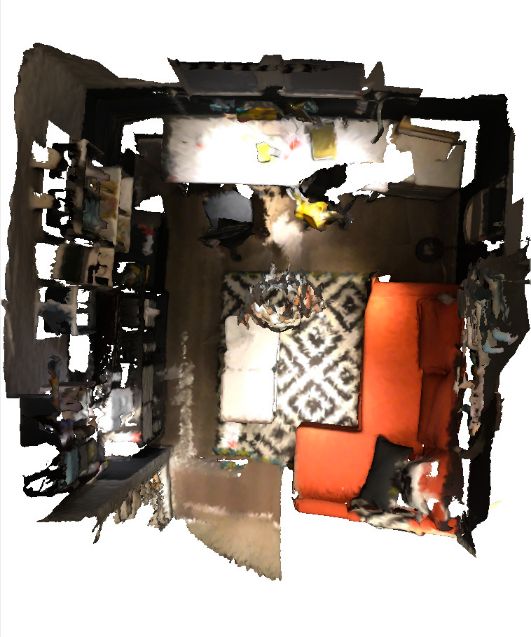} &
    \includegraphics[width=\sz\textwidth]{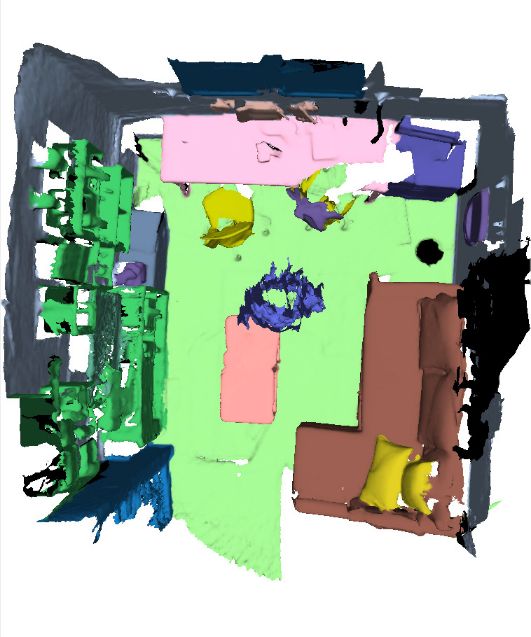} & 
    \includegraphics[width=\sz\textwidth]{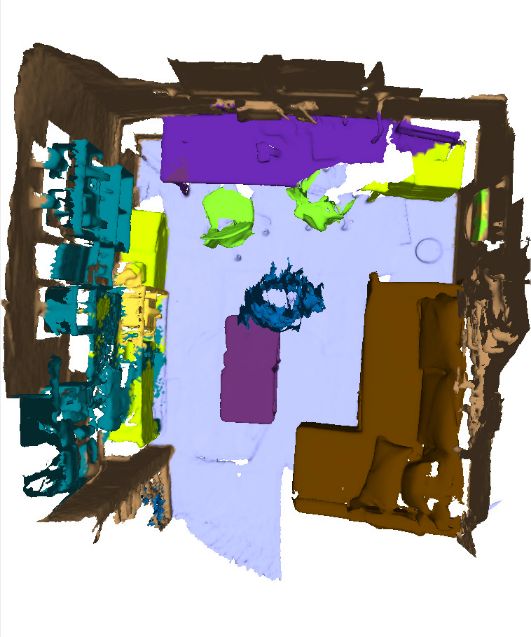} &
    \includegraphics[width=\sz\textwidth]{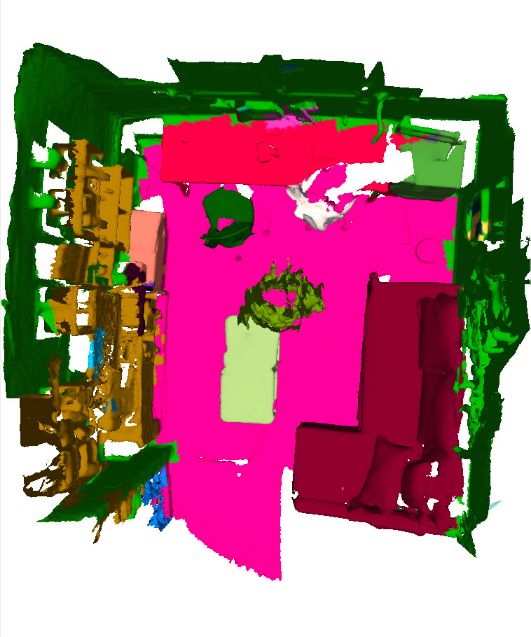} & 
    \includegraphics[width=\sz\textwidth]{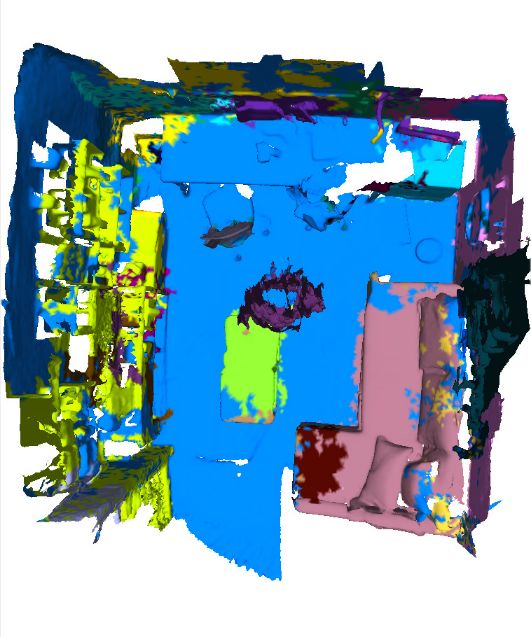} & 
    \includegraphics[width=\sz\textwidth]{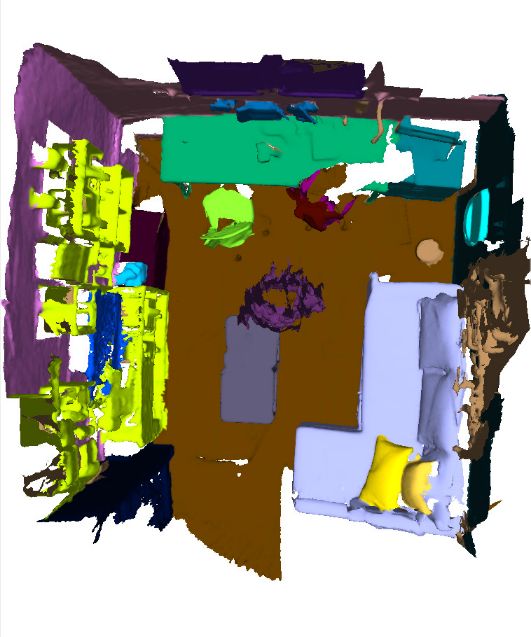} & 
    \includegraphics[width=\sz\textwidth]{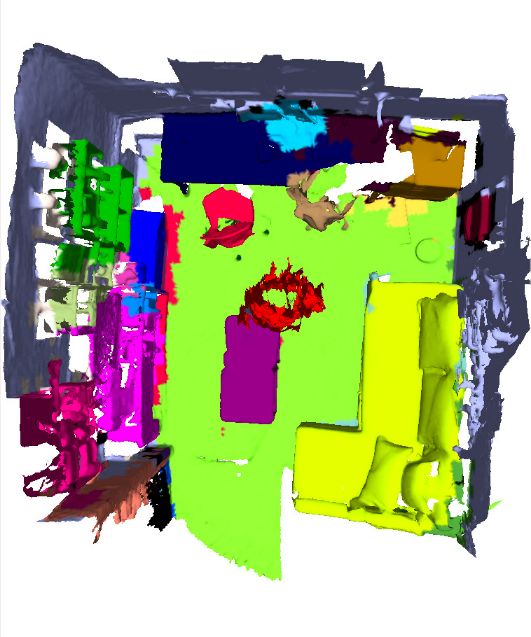} \\[-0.2cm]
    Input (RGB) & Semantic GT &  SPC~\cite{3DSemanticSegmentationWithSubmanifoldSparseConvNet} & CC & SGPN~\cite{Wang-et-al-CVPR-2018} & Instance GT & Ours \\
  \end{tabular}
  \vspace{0.005cm}
  \caption{\textbf{Qualitative results of our method on the ScanNet validation dataset~\cite{Dai-et-al-CVPR-2017}.}
  This figure shows the original input scene as a textured mesh, the semantic labeling results of SparseConvNet (SPC)~\cite{3DSemanticSegmentationWithSubmanifoldSparseConvNet} which we use as input and our instance labeling results as well as the semantic groundtruth (GT). We further show multiple 3D instance segmentation baselines: connected component (CC) labeling on the SPC semantic labeling, SPGN~\cite{Wang-et-al-CVPR-2018}, and the groundtruth instance labels next to our labeling results.}
  \label{fig:qualitative_results_a}
\end{figure*}

Table~\ref{tab:results_test} provides an overview of our benchmark results on the ScanNet test dataset (with held out ground truth). One can see that our method outperforms the others in AP50 score. Other methods include those that process all RGB-D images that were used to reconstruct the scenes of ScanNet. Instance labels of single RGB-D frames in these methods are propagated throughout the whole scene and concatenated based on the location estimation. On the other hand, our method directly operates in the 3D setting, without the need to use the 2D information. This leads to much faster processing on the 3D scenes, and requires substantially less information to extract the 3D object instance segmentations.

\section{Conclusion}
We proposed a method for 3D instance segmentation of voxel-based scenes.
Our approach is based on metric learning and the first part assigns all voxels belonging to the same object instance feature vectors that are in close vicinity. 
Conversely, voxels belonging to different object instances are assigned  features that are further apart from each other in the feature space. 
The second part estimates directional information of object centers, which is used to score the segmentation results generated by the first part.

{\small
\vspace{0.5em}
\boldparagraph{Acknowledgments.}
This research was supported by competitive funding from King Abdullah University of Science and Technology (KAUST). Further support was received by the Intelligence Advanced Research Projects Activity (IARPA) via Department of Interior/ Interior Business Center (DOI/IBC) contract number D17PC00280.
The U.S. Government is authorized to reproduce and distribute reprints for Governmental purposes notwithstanding any copyright annotation thereon. Disclaimer: The views and conclusions contained herein are those of the authors and should not be interpreted as necessarily representing the official policies or endorsements, either expressed or implied, of IARPA, DOI/IBC, or the U.S. Government.
\par
}


{\small
\bibliographystyle{ieee_fullname}
\bibliography{bibliography}
}

\end{document}